\documentclass[11pt]{article}

\usepackage[final]{acl}

\usepackage{times}
\usepackage{latexsym}

\usepackage[T1]{fontenc}

\usepackage[utf8]{inputenc}

\usepackage{microtype}

\usepackage{inconsolata}

\usepackage{graphicx}
\graphicspath{{figures/}} 

\usepackage{times}
\usepackage{latexsym}
\usepackage{amsmath}
\usepackage{xcolor}
\usepackage{url}
\usepackage{booktabs}
\usepackage{microtype}
\usepackage{multirow}

\usepackage{subcaption} 
\usepackage{adjustbox}
\usepackage[table]{xcolor}
\usepackage{longtable,booktabs}
\usepackage{float}                 
\usepackage[section]{placeins}     
\usepackage{dblfloatfix}           
\usepackage{dsfont}
\usepackage{hyperref}

\title{Systematic Evaluation of Uncertainty Estimation Methods in Large Language Models}

\author{Christian Hobelsberger\thanks{~Equal contribution.\\ Contact emails: Ch.Hobelsberger@campus.lmu.de, the.winner@campus.lmu.de} \\
\textit{LMU Munich, Munich Re, relAI} \\
\And
Theresa Winner\footnotemark[1] \\
\textit{LMU Munich} \\
\And
Andreas Nawroth \\
\textit{Munich Re} \\
\AND
Oliver Mitevski \\
\textit{Munich Re} \\
\And
Anna-Carolina Haensch \\
\textit{LMU Munich, MCML, University of Maryland}}

\begin{document}
\maketitle

\begin{abstract}
Large language models (LLMs) produce outputs with varying levels of uncertainty, and, just as often, varying levels of correctness; making their practical reliability far from guaranteed. 
To quantify this uncertainty, we systematically evaluate four approaches for confidence estimation in LLM outputs:
\textbf{VCE}, \textbf{MSP}, \textbf{Sample Consistency}, and \textbf{CoCoA} \citep{Vashurin2025b}.

For the evaluation of the approaches, we conduct experiments on four question-answering tasks using a state-of-the-art open-source LLM. 
Our results show that each uncertainty metric captures a different facet of model confidence and that the hybrid CoCoA approach yields the best reliability overall, improving both calibration and discrimination of correct answers. 
We discuss the trade-offs of each method and provide recommendations for selecting uncertainty measures in LLM applications.
\end{abstract}

\section{Introduction}
Large language models (LLMs) and their outputs can be difficult to trust: models may produce incorrect answers with unwarranted confidence or hedge on easy questions. 
This unreliability is especially problematic in high-stakes applications where undetected errors can have serious consequences. 
One way to mitigate this risk is to equip LLMs with \emph{uncertainty estimation} mechanisms that quantify the model's confidence in its own outputs. 
Accurate confidence estimates enable downstream systems to calibrate predictions, abstain from low-confidence outputs, or request human intervention when needed \citep{Ji2023}. 

Since uncertainty can arise from different sources that affect model behavior, prior work distinguishes between \emph{aleatoric} uncertainty, arising from inherent data noise, and \emph{epistemic} uncertainty, reflecting limited model knowledge \citep{DerKiureghian2009}. 
Estimating these uncertainties in LLMs is challenging because their predictions are text tokens. 
Recent studies propose proxy metrics, including model probability distributions \citep{Hendrycks2017, Jiang2021b, Chen2023}, elicited self-reports \citep{Kadavath2022, Xiong2023}, and multi-sample consistency signals \citep{Wang2022, Lyu2024, Wang2024}. 
Each has advantages and limitations, and it remains unclear which methods are most effective across tasks.

We compare \textbf{Verbalized Confidence Elicitation} (\textsc{VCE}) asks the model to state its own confidence; \textbf{Maximum Sequence Probability} (\textsc{MSP}) derives confidence from the probability assigned to the output; \textbf{Sample Consistency} measures uncertainty through the stability of answers across multiple sampled outputs, and \textbf{Confidence-Consistency Aggregation} (\textsc{CoCoA}) combines model-internal confidence with output consistency. 

We make four main contributions: (1) we  provide a unified implementation and benchmarking of four uncertainty estimation methods for LLMs, including a hybrid that fuses confidence and consistency \citep{Vashurin2025b}; 
(2) we introduce compact calibration via confidence-like scores;
(3) we develop a multi-dataset evaluation of calibration and discrimination; and 
(4) we provide practical guidance for deploying uncertainty measures.

\section{Related Work}

\paragraph{Calibration and confidence for NLP/LLMs.}
Modern neural networks, including transformers, are often miscalibrated, yielding overconfident predictions even when incorrect \citep{Guo2017, Kumar2019}. This miscalibration undermines reliability in downstream applications, since confidence scores no longer reflect true likelihoods of correctness \citep{Hendrycks2017, Lakshminarayanan2017}. Post-hoc methods can partially mitigate these issues \citep{Desai2020}. For generative models, token or sequence likelihoods can be informative, since higher probabilities often signal model confidence while lower probabilities reflect ambiguity, though these measures remain sensitive to length and decoding choices \citep{Ott2018}. LLMs can self-assess to some degree, for example by assigning confidence scores to their own answers or expressing subjective probabilities in natural language \citep{Kadavath2022}, though raw self-reports tend to be overconfident \citep{Xiong2023}.

\paragraph{Consistency and multi-sample reasoning.}
Consistency-based approaches, including self-consistency via multiple sampled chains, have shown promising results for detecting uncertainty and hallucinations in black-box LLMs \citep{Wang2023SelfConsistency, Manakul2023SelfCheckGPT}. Recent hybrid methods, such as the CoCoA approach, combine confidence from likelihoods with agreement signals, improving calibration and discrimination \citep{Vashurin2025b}.

\paragraph{Selective prediction and abstention.}
Rather than forcing a prediction for every case, models can be trained to respond only when confident and abstain otherwise. This lowers coverage but improves the reliability of the outputs, making real-world deployment more trustworthy \citep{Geifman2017Selective,Geifman2019SelectiveNet}.


\section{Uncertainty Estimation Methods}
We consider four approaches for estimating the uncertainty of an LLM's output on a given query. 
Each produces a confidence or uncertainty score that can be compared against the correctness of the model's answer.

\subsection{Verbalized Confidence Elicitation (VCE)}
A simple approach is to ask the model for its own confidence: Verbalized Confidence Elicitation (\textsc{VCE}) prompts it to provide both an answer and a self-reported confidence score \citep{Xiong2023}.
For example, the model might be prompted with: \emph{``Answer the question and provide your confidence (0--100).''} 
A possible response could be: \emph{``Answer: 42. Confidence: 80.''} 
This approach is model-agnostic (requiring no access to model internals) and easy to implement. 

Prior work suggests that LLMs can sometimes provide reasonably calibrated self-evaluations when prompted appropriately \citep{Kadavath2022, Geng2023}. However, in practice, we find that verbalized confidence tends to be systematically overestimated, particularly for instruction-tuned models that are trained to sound assured \citep{Wang2024, Kalai2025}.
Models often express high confidence even when they are incorrect, limiting the utility of raw VCE as a reliability signal. 
We also experiment with using \emph{multiple} samples in the VCE setting, inspired by \citet{Xiong2023}. 
Specifically, we obtain $M$ independent answers (via repeated sampling), each with a self-reported confidence $C_i$. 
We then compute an \emph{agreement-weighted confidence} that gives higher weight to answers that agree with the majority prediction:
\begin{equation}
    \hat{C}_{\text{agree}} \;=\; \frac{\sum_{i=1}^{M} \mathds{1}\{\hat{Y}_i = \tilde{Y}\} \cdot C_i}{\sum_{i=1}^{M} C_i}\,,
\end{equation}
where $\tilde{Y}$ is the answer given by the majority of samples and $\mathds{1}\{\cdot\}$ is an indicator for agreement. 
This aggregated confidence $\hat{C}_{\text{agree}}$ is used as the final VCE score for that query. 
Intuitively, it discounts confident outlier answers that disagree with the majority, which helps reduce overconfidence in the combined estimate.

\subsection{Maximum Sequence Probability (MSP)}
\label{sec:msp}
Probability-based metrics derive uncertainty from the model's internal likelihood assigned to its output. 
A simple and widely used measure is \textsc{Maximum Sequence Probability} (\textsc{MSP}), analogous to the Maximum Softmax Probability for classifiers \citep{Hendrycks2017}. 
Given an input $x$ and an output sequence $y = [y_1, y_2, \dots, y_T]$ produced by the model, we define the \emph{unnormalized uncertainty} as the negative log-likelihood of the sequence:
\begin{equation}
    U_{\text{MSP}}(x,y) \;=\; -\sum_{t=1}^{T} \log P(y_t \mid y_{<t},\,x)\,. 
\end{equation}
A lower $U_{\text{MSP}}$ (higher probability) indicates higher model confidence in that particular output. 
MSP can be obtained with negligible overhead during generation, since modern decoding APIs often return token probabilities.

Direct values of $U_{\text{MSP}}$ are not comparable across different inputs or tasks because they depend on sequence length and other factors (see Appendix Figure \ref{fig:msp_uncertainty}). 
Extending prior work, we apply a normalization to convert $U_{\text{MSP}}$ into a bounded \emph{confidence-like score} $C_{\text{MSP}}(x,y) \in [0,1]$ \citep{Hendrycks2017, Jiang2021b}. 
We first clip extreme values by truncating $U_{\text{MSP}}$ at the 98th percentile over the evaluation set. This mitiagates outliers. 
Denote this clipped uncertainty as $\tilde{U}$. 
We then apply min-max normalization 
\begin{equation}
    C_{\text{MSP}}(x,y) \;=\; 1 \;-\; \frac{\tilde{U}(x,y) - \min(\tilde{U})}{q_{0.98} - \min(\tilde{U})}\,,
\end{equation}
where $q_{0.98}$ is the 98th percentile of $U_{\text{MSP}}$ values (after clipping) and $\min(\tilde{U})$ is the minimum. 
Thus, $C_{\text{MSP}}$ maps the model's likelihood for an output onto a $[0,1]$ scale, where 1 corresponds to high confidence. 
Although this normalization makes scores more comparable, MSP-based confidence still suffers from known issues: models tend to assign overly high probability to incorrect answers \citep{Guo2017, Jiang2021a}, and longer outputs incur lower probabilities by construction \citep{Ott2018}. 
These limitations motivate augmenting probability-based measures with additional signals.

\subsection{Sample Consistency}
\label{sec:cons}
Another perspective on uncertainty is to check if an LLM produces the same answer when asked the same question multiple times. 
\textbf{Sample consistency} assesses the stability of model outputs under stochastic generation: if the model almost always gives the same answer, it is likely confident; if it frequently changes its answer, it is uncertain. 
This method captures a form of epistemic uncertainty by probing the model's decision boundary with multiple samples \citep{Lyu2024}. 

In our experiments, we generate $k$ answers for each question using either (a) repeated decoding with different random seeds, or (b) requesting multiple outputs in one call (e.g., top-$k$ sampling). 
We then measure pairwise semantic similarity among the $k$ answers. 
High agreement indicates higher confidence, whereas diverse or contradictory answers signal uncertainty. 
We use two complementary similarity measures:
\begin{itemize}\setlength\itemsep{0em}
    \item \textbf{Embedding Similarity:} We compute the cosine similarity between each pair of answers using Sentence-BERT embeddings \citep{Reimers2019}. This captures semantic closeness (e.g., paraphrases).
    \item \textbf{NLI-Based Consistency:} We also evaluate whether one answer entails the other using a RoBERTa-large model fine-tuned on natural language inference (NLI) \citep{Liu2019, Williams2018}. For each pair, we take the predicted probability of \emph{entailment} as a directional consistency score. This detects contradictions that cosine similarity might miss.
\end{itemize}
For each question, we aggregate the pairwise similarities across all $M=\binom{k}{2}$ pairs:
\begin{equation}
    U_{cons}(x)=\frac{1}{M} \sum_{i<j} sim(x_i, x_j)
\end{equation}

The average similarity provides a \emph{consistency score} for that question, while the variance or minimum similarity can indicate divergence among outputs. In practice, we found the mean similarity to be a simple and effective proxy for consistency, which we use as the Sample Consistency metric.

Sample consistency directly captures whether the model's knowledge and reasoning are robust for a given query \citep{Wang2022}. It can reveal uncertainty even when probability-based metrics are overconfident, and it does not require ground-truth answers or model introspection. However, it comes at a high computational cost (multiple model runs) and can be fooled if the model consistently repeats the \emph{same wrong answer}. In such cases, consistency alone would be high even though the model is confidently wrong.

\subsection{Confidence--Consistency Aggregation (CoCoA)}
\label{sec:cocoa}
Finally, we evaluate \textbf{CoCoA} \citep{Vashurin2025b}, a recently proposed method that combines the model's internal confidence with output consistency. 
CoCoA is formulated in a minimum Bayes risk framework: it penalizes an answer $y^*$ if (i) the model assigns low probability to $y^*$ (information-based uncertainty $u$), and (ii) $y^*$ is semantically dissimilar to other likely answers (consistency-based uncertainty $U_{\text{cons}}$). 
In our implementation we instantiate the first term with sequence likelihood, \(u(y^\star\!\mid x) = -\log P(y^\star\!\mid x)\), using the same token log-probabilities as MSP (Sec.~\ref{sec:msp}). 
\emph{While \(u\) could also be instantiated with VCE, we prefer MSP because it is prompt-agnostic, numerically stable across tasks, and less overconfident in our setting (Sec.~\ref{sec:results}), which leads to more reliable fusion with the consistency term.} 
For the consistency term, we generate $M$ alternative answers $\{y^{(i)}\}_{i=1}^M$ via stochastic decoding and compute their semantic similarity to $y^*$ using the same NLI-based model as for Sample Consistency (see Chapter \ref{sec:cons}). 
Let $s(y^*, y') \in [0,1]$ be the similarity (entailment probability) between $y^*$ and another output $y'$. 
The average dissimilarity is:
\begin{equation}
    U_{\text{cons}}(y^*|x) \;=\; \frac{1}{M}\sum_{i=1}^{M} \big(1 - s(y^*,\,y^{(i)})\big)\,. 
\end{equation}
CoCoA combines the two components multiplicatively:
\begin{equation}
    \hat{U}_{\text{CoCoA}}(y^*|x) \;=\; u(y^*|x) \;\cdot\; U_{\text{cons}}(y^*|x)\,,
\end{equation}
such that high uncertainty is only assigned when the model is both internally unsure and externally inconsistent with itself. 
The raw $\hat{U}_{\text{CoCoA}}$ values are then mapped to a \textit{confidence-like score} $C_{CoCoA}$ in $[0,1]$ via the same clipping and normalization procedure used for MSP (applying it to $\hat{U}_{\text{CoCoA}}$ across the dataset, see Chapter \ref{sec:msp}). 
Thus, CoCoA produces a final confidence-like score $C_{CoCoA} = 1 - \tilde{U}_{\text{CoCoA,norm}}$ for each answer.

CoCoA effectively fuses two signals: the model's own confidence in an answer and the semantic relationship between multiple outputs generated using repeated sampling
Prior evaluations have shown that CoCoA can improve correlation with correctness over using either confidence or consistency alone \citep{Vashurin2025b}.
The main drawback is the computational overhead of generating multiple samples for each query. In settings where latency is critical, CoCoA might be too expensive, but it represents a strong upper bound on what post-hoc uncertainty quantification can achieve by leveraging all available signals. 
\emph{For high-risk deployments, a more conservative approach to fusion is also possible, such as an OR-style rule }$U_{\text{OR}}(y^\star\!\mid x)=\max\{u(y^\star\!\mid x),\,U_{\text{cons}}(y^\star\!\mid x)\}$. \emph{This tends to reduce coverage, but also lowers the likelihood of accepting uncertain answers.}

\paragraph{Resources and tooling.} We use Sentence-BERT \citep{Reimers2019} and RoBERTa \citep{Liu2019} with MultiNLI \citep{Williams2018}, and build on HuggingFace Transformers \citep{Wolf2020}. For the evaluation, we use MSP and CoCoA (see Section \ref{sec:msp} and \ref{sec:cocoa}, respectively) using the \texttt{lm-polygraph} package \citep{IINemo2025}. Tasks include BoolQ \citep{Clark2019}, SQuAD\,2.0 \citep{Rajpurkar2018}, TriviaQA \citep{Joshi2017}, and GSM8K \citep{Cobbe2021}. Base models follow the Llama\,3 family \citep{Llama3-2024}. 

\section{Experimental Setup}

\subsection{Tasks and Model}
Our goal is to stress-test uncertainty estimators across qualitatively different cases:
(i) \emph{binary} decisions with short answers, where overconfidence is common; (ii) \emph{extractive} reading comprehension tasks with long context passages; (iii) \emph{open-domain} factoid QA with knowledge gaps and entity confusion; and (iv) \emph{multi-step reasoning} with arithmetic and intermediate steps. To that end, we evaluate on four standard QA benchmarks:

\paragraph{BoolQ} \citep{Clark2019} is a yes/no dataset derived from natural queries. Answers are single tokens (\textsc{yes}/\textsc{no}), providing a straightforward testbed for calibration, since probabilities directly correspond to the correctness of yes/no decisions. Ambiguity in question phrasing induces aleatoric uncertainty, while topical drift can elicit epistemic uncertainty.

\paragraph{SQuAD\,2.0} \citep{Rajpurkar2018} augments extractive QA with unanswerable questions. Models must either return a short span from the supplied passage or abstain, making it ideal for studying selective prediction and calibration. 

\paragraph{TriviaQA} \citep{Joshi2017} is an open-domain factoid QA. Evidence passages are not guaranteed to contain a succinct rationale, and entity-heavy questions surface knowledge gaps and plausible-but-wrong confusions. This setting is known to benefit ranking-style confidence (e.g., likelihood-based scores) while remaining challenging to calibrate.

\paragraph{GSM8K} \citep{Cobbe2021} contains grade-school math word problems that require multi-step reasoning and arithmetic. Even small reasoning slips lead to confidently wrong final numbers, which stresses both discrimination (AUROC) and calibration. We ask the model for a short final answer; intermediate steps may be produced but are not required.

\paragraph{Model.}
We use \textbf{LLaMA~3.2 (3B-Instruct)} \citep{Meta2025,Llama3-2024} for three reasons. 
First, it is \emph{open} and exposes token log-probabilities, enabling MSP without internal modifications and making our pipeline reproducible. 
Second, the \emph{3B} size offers a strong accuracy and compute trade-off: it is capable enough to make the uncertainty problem non-trivial while being small enough that we can afford repeated sampling (SEP) at scale. 
Third, the instruction-tuned variant produces structured self-reports for \textsc{VCE} and handles long contexts (e.g., SQuAD passages) with its large context window. 
While larger LLaMA variants may yield higher raw accuracy, our focus is on the \emph{relative behavior} of uncertainty estimators; we expect the qualitative trends we observe (e.g., VCE overconfidence, gains from consistency, CoCoA’s calibration) to transfer across model sizes.

\subsection{Decoding \& multi-sample regimes}
We adopt a temperature of $T{=}0.7$ for all uncertainty estimation methods, as it provides a robust trade-off across tasks (see Figs.~\ref{fig:temp_suite_single} and \ref{fig:temp_suite_multi}). 
For \emph{multi-sample} conditions ($M{=}5$ unless noted), we evaluate two regimes \emph{for VCE and Sample Consitency}: 
\textbf{TOP-K} (multi-output from one decoding call) and 
\textbf{SEP} (five separate decodings with independent seeds). 
For \emph{CoCoA} we \textbf{always use SEP} to obtain $M{=}10$ independent samples; TOP-K is not used.\\
The resulting $M$ answers are aggregated per method (agreement-weighted confidence for VCE, pairwise similarity for consistency, and the combined confidence-consistency score for CoCoA). 

\paragraph{Effective sample sizes ($N$) and filtering.}
\label{sample_size_N}
Unless noted otherwise, we target $N{=}1000$ examples per dataset for \emph{single-sample VCE} and $N{=}500$ for all other methods (including multi-sample VCE, Sample Consistency, MSP, and CoCoA). 
The realized $N$ can be slightly smaller due to automatic filtering of unparseable model outputs (e.g., missing or non-numeric VCE self-reports, malformed JSON), empty/overlong generations, or dataset-specific preprocessing.
For SQuAD\,2.0, we remove unanswerable questions to align with our evaluation prompts, yielding $N{=}638$ for the single-sample condition and $N{=}310$ in the multi-sample setting.

\subsection{Evaluation Metrics}
To quantify the effectiveness of each uncertainty estimation method, we consider two primary evaluation criteria: \textbf{Calibration} and \textbf{Discrimination}. 
Calibration measures how well the confidence scores align with actual correctness frequencies, while discrimination measures how well the scores separate correct vs incorrect outputs.

\textbf{Expected Calibration Error (ECE).} We compute the expected calibration error (ECE) \citep{Naeini2015} of each method's confidence scores. 
ECE partitions the predictions into bins by confidence and compares the average confidence to the empirical accuracy in each bin. 
If $B_b$ is the set of indices of predictions falling into bin $b$, the ECE is:
\begin{equation}
    \text{ECE} = \sum_{b=1}^{B} \frac{|B_b|}{n} \big| \text{acc}(B_b) - \text{conf}(B_b)\big|\,,
\end{equation}
where $\text{acc}(B_b)$ is the fraction of correct answers in bin $b$ and $\text{conf}(B_b)$ is the average predicted confidence in that bin, and $n$ is the total number of predictions. 
Lower ECE (closer to 0) indicates better calibration (the confidence scores reflect true probabilities of correctness). We use $B=10$ bins of equal width over $[0,1]$.

\textbf{AUROC.} To evaluate discrimination, we use the Area Under the Receiver Operating Characteristic curve (AUROC) \citep{Fawcett2006}. 
This treats confidence-based ranking as a binary classification of correct vs. incorrect answers. 
The ROC curve plots true positive rate vs false positive rate as we vary a confidence threshold $\tau$. 
The AUROC is the probability that a randomly chosen correct answer is assigned a higher confidence than a randomly chosen incorrect answer. 
An AUROC of 1.0 indicates perfect rank ordering by confidence, whereas 0.5 is no better than chance. 
We calculate AUROC for each method by comparing its confidence scores against the ground-truth correctness of the model's answers.

Additionally, we report the \textbf{accuracy} (fraction of correct answers) of the model on each dataset to contextualize the difficulty and the headroom for uncertainty metrics (which cannot exceed the model's own accuracy). 
Note that methods like CoCoA and MSP do not change which answer is given (they only score the given answer), so accuracy remains the same across those methods.

\section{Results and Analysis}\label{sec:results}
Table~\ref{tab:results_overview} summarizes performance across datasets. 
In addition to \textbf{Accuracy}, \textbf{ECE}, and \textbf{AUROC}, we report \textbf{Average Confidence} (in \%) to expose over/under-confidence directly.
\emph{Sample sizes} ($N$, rightmost column) follow the filtering rules in Section~\ref{sample_size_N}: minor deviations occur for VCE due to the automatic removal of unparseable generations.
For SQuAD\,2.0, unanswerable questions were removed, yielding $N{=}638$ in the single-sample case and $N{=}310$ in the multi-sample setting.

\begin{table*}[!t]
\centering
\footnotesize
\setlength{\tabcolsep}{6pt}
\begin{adjustbox}{max width=\textwidth}
\begin{tabular}{llccccc}
\toprule
\textbf{Dataset} & \textbf{Method} & \textbf{Accuracy $\uparrow$} & \textbf{Avg. Con-} & \textbf{ECE $\downarrow$} & \textbf{AUROC $\uparrow$} & \textbf{N} \\
\textbf{} & \textbf{} & \textbf{} & \textbf{fidence (\%)} & \textbf{} & \textbf{} & \textbf{} \\
\midrule

\multirow{6}{*}{\textbf{BoolQ}} 
& VCE: Single Sample & 0.780 & 90.371 & 0.183 & 0.646 & 973 \\
& VCE: Multi-Sample (TOP-K) & 0.768 & 88.925 & \textbf{0.146} & 0.646 & 500\\
& VCE: Multi-Sample (SEP) & 0.768 & 91.107 & 0.168 & 0.636 & 500 \\
& MSP & \textbf{0.790} & 58.714 & 0.203 & 0.673 & 500 \\
& CoCoA & \textbf{0.790} & 65.219 & 0.160 & \textbf{0.687} & 500 \\
\midrule

\multirow{8}{*}{\textbf{SQuAD}} 
& VCE: Single Sample & 0.868 & 97.633 & 0.124 & 0.573 & 638 \\
& VCE: Multi-Sample (TOP-K) & 0.829 & 77.338 & 0.118 & 0.722 & 310 \\
& VCE: Multi-Sample (SEP) & 0.848 & 77.372 & 0.113 & 0.746 & 310\\
& Sample Cons. (TOP-K) & 0.865 & 61.616 & 0.186 & 0.739 & 310\\
& Sample Consistency (SEP) & 0.832 & 61.506 & 0.149 & 0.758 & 310\\
& MSP & \textbf{0.890} & 78.023 & 0.122 & 0.836 & 310\\
& CoCoA & \textbf{0.890} & 83.229 & \textbf{0.062} & \textbf{0.844} & 310 \\
\midrule

\multirow{8}{*}{\textbf{TriviaQA}} 
& VCE: Single Sample & 0.501 & 68.123 & 0.309 & 0.700 & 994 \\
& VCE: Multi-Sample (TOP-K) & 0.511 & 70.564 & 0.216 & 0.645 & 489 \\
& VCE: Multi-Sample (SEP) & 0.496 & 71.456 & 0.233 & 0.690 & 500 \\
& Sample Cons. (TOP-K) & 0.508 & 37.491 & 0.162 & 0.793 & 500\\
& Sample Consistency (SEP) & 0.510 & 38.122 & \textbf{0.134} & 0.793 & 500\\
& MSP & \textbf{0.610} & 65.186 & 0.147 & \textbf{0.841} & 500 \\
& CoCoA & 0.608 & 68.662 & 0.167 & \textbf{0.841} & 500 \\
\midrule

\multirow{5}{*}{\textbf{GSM8K}} 
& VCE: Single Sample & 0.651 & 98.800 & 0.335 & 0.548 & 988 \\
& VCE: Multi-Sample (TOP-K) & 0.749 & 46.800 & 0.129 & 0.775 & 462 \\
& VCE: Multi-Sample (SEP) & \textbf{0.781} & 47.000 & 0.145 & 0.736 & 468 \\
& MSP & 0.690 & 64.557 & 0.104 & \textbf{0.838} & 500 \\
& CoCoA & 0.680 & 65.219 & \textbf{0.081} & 0.786 & 500 \\
\bottomrule
\end{tabular}
\end{adjustbox}
\caption{Accuracy, calibration (ECE), and discrimination (AUROC) on each dataset, plus average confidence. 
Multi-sample \emph{VCE} and Sample Consistency uses $M{=}5$ with either TOP-K (single decoding with multiple returns) or SEP (five separate decodings). 
CoCoA use $M{=}10$ via SEP only. 
Bold = best (lowest ECE or highest AUROC) per dataset.}
\label{tab:results_overview}
\vspace{-0.5em}
\end{table*}

\begin{figure*}[t]
  \centering
  \includegraphics[width=\textwidth]{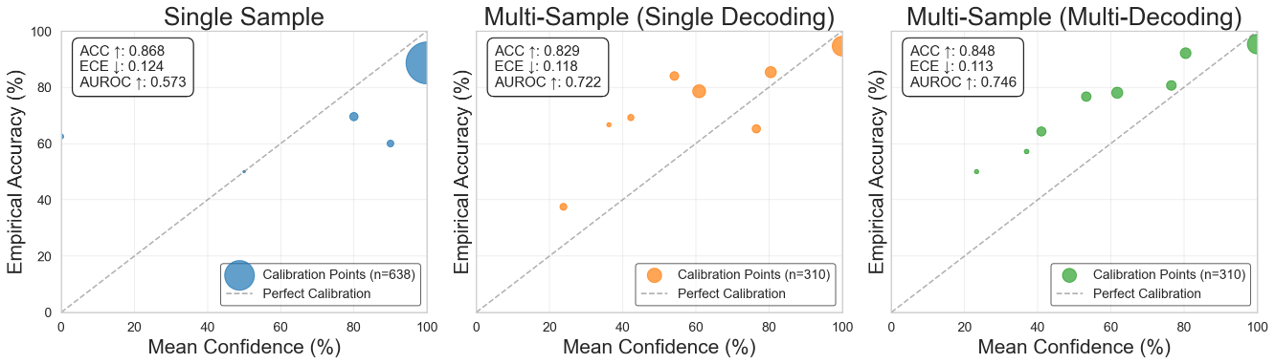}
  \caption{\textbf{VCE calibration on SQuAD\,2.0 across decoding methods.}
  Reliability plots (empirical accuracy vs.\ mean confidence) for single-sample,
  multi-sample with single decoding (TOP-K), and multi-sample with multi-decoding (SEP).
  Bubble area is proportional to bin count; the dashed line denotes perfect calibration $y{=}x$.
  Multi-decoding tightens high-confidence bins, reduces ECE, and improves AUROC while keeping accuracy comparable.}
  \label{fig:VCE_squad_enlarged_fonts}
\end{figure*}

\paragraph{High-level takeaways.}
(1) \textbf{VCE overconfidence is pronounced.} Single-sample VCE reports extremely high mean confidence (e.g., $97.6\%$ on SQuAD, $98.8\%$ on GSM8K) despite much lower accuracy, yielding large ECE. 
(2) \textbf{Multi-sample aggregation moderates VCE.} Aggregating self-reports reduces average confidence by 20–50pp and improves calibration/AUROC (e.g., BoolQ ECE $0.183$ $\to$ \textbf{$0.146$}; SQuAD AUROC $0.573 \to 0.722$–$0.746$). 
(3) \textbf{Hybridization pays off.} CoCoA is best-calibrated (SQuAD \textbf{$0.062$}, GSM8K \textbf{$0.081$}) with competitive AUROC (SQuAD \textbf{$0.844$}; tie on TriviaQA).

\paragraph{Dataset-level analysis (task difficulty matters).}
\emph{BoolQ} (binary yes/no) is comparatively easy; all methods achieve similar accuracy ($\approx0.79$), and CoCoA provides the best ranking (AUROC \textbf{$0.687$}) while VCE(TOP-K) attains the lowest ECE (\textbf{$0.146$}). 
\emph{SQuAD\,2.0} (extractive QA with unanswerables) benefits from combining confidence and consistency: CoCoA achieves both the lowest ECE and highest AUROC at top accuracy ($0.890$). 
\emph{TriviaQA} (open-domain factoid) stresses retrieval/knowledge; MSP ranks best (\textbf{$0.841$}), while \emph{consistency (SEP)} calibrates best (\textbf{$0.134$}). 
\emph{GSM8K} (multi-step math reasoning) is the hardest: self-consistency boosts accuracy (VCE-SEP \textbf{0.781}); MSP remains the best ranker (\textbf{$0.838$}); CoCoA is clearly the most calibrated (\textbf{$0.081$}).

\paragraph{TOP-K vs.\ SEP.}
Across datasets, TOP-K and SEP are \emph{largely comparable}. For VCE multi-sample, the mean absolute difference between TOP-K and SEP is $\approx0.03$ in AUROC and $\approx0.015$ in ECE (e.g., BoolQ AUROC $0.646$ vs.\ $0.636$; SQuAD $0.722$ vs.\ $0.746$; TriviaQA $0.645$ vs.\ $0.690$; GSM8K $0.775$ vs.\ $0.736$). Accuracy differences average $\approx0.02$ (e.g., GSM8K $0.749$ vs.\ $0.781$). 
Sample Consistency with separated decoding (SEP) achieves slightly higher AUROC values than Top-K on SQuAD ($0.758$ vs. $0.739$) and the same in TriviaQA ($0.794$). Top-K consistently achieves lower ECE values($0.186$ vs. $0.149$ on SQuAD; $0.162$ vs. $0.134$ on TriviaQA), suggesting that it provides better calibrated confidence scores.
This near parity yields a practical recommendation: \textbf{when compute/latency matters, prefer TOP-K} (single decoding, multiple returns), using SEP only when small diversity gains are needed. 
(For CoCoA we use SEP exclusively.)

As suggested by Figure~\ref{fig:VCE_squad_enlarged_fonts}, the dominant failure mode for single-sample VCE is systematic overconfidence, which multi-sample aggregation largely mitigates.
Calibration plots for all datasets are provided in Appendix~Fig.~\ref{fig:vce_methods}.

\begin{figure}[H]
    \centering
    \includegraphics[width=\linewidth]{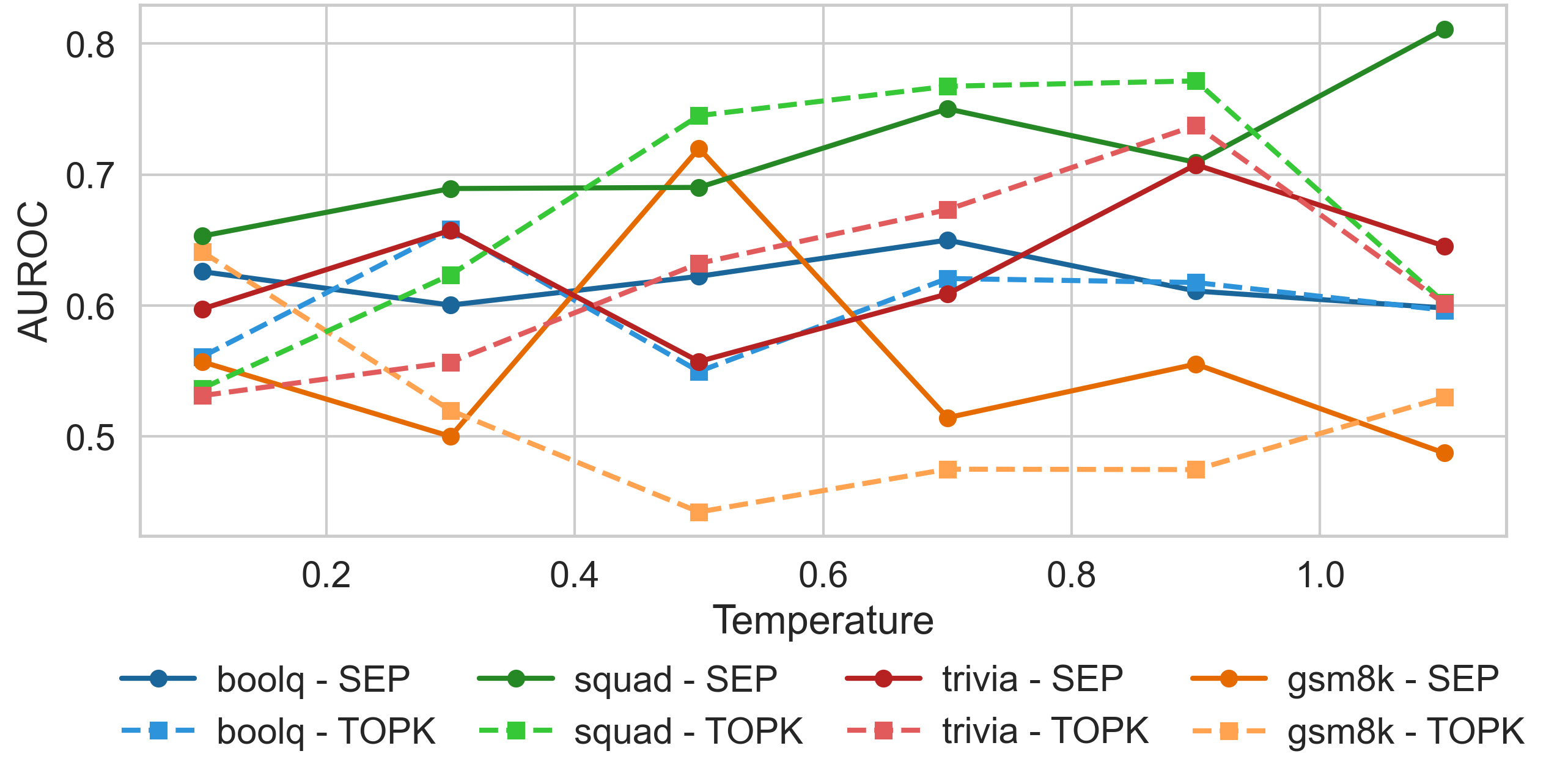}
    \caption{\textbf{Temperature effects on ranking quality.} 
    AUROC vs.\ decoding temperature for multi-sample VCE under SEP (solid) and TOP-K (dashed).
    Optima differ by dataset, with most gains occurring at mid-range $T$ (0.6–0.8).}
    \label{fig:temp_multi_auroc}
\end{figure}

\paragraph{Temperature sweeps: no universal best.}
We conducted comprehensive temperature sweeps evaluating AUROC, ECE, and overconfidence across all tasks.
Figure~\ref{fig:temp_multi_auroc} shows one representative result: AUROC vs.\ temperature under multi-sample VCE. This figure illustrates that temperature effects are \emph{dataset-specific}.
AUROC typically increases from very low $T$ ($0.1$–$0.3$) toward mid-range values ($\approx0.7$), though the precise optimum varies across datasets.
At higher temperatures, calibration error and overconfidence often plateau or worsen.
\emph{There is no single temperature that is uniformly optimal across tasks and evaluation metrics}; $T$ must be adjusted for each dataset and objective.
The full single-sample and multi-sample sweeps, including accuracy, calibration error, and bias, are provided in Appendix Figures \ref{fig:temp_suite_single} and \ref{fig:temp_suite_multi}

\paragraph{MSP.}
\textsc{MSP} achieves strong discrimination with the highest AUROC values on TriviaQA ($0.841$) and GSM8K ($0.838$), making it a dependable ranker.
However, calibration is slightly weaker in some cases. Since ECE is typically higher than CoCoA's, the probability scores are less aligned with true correctness (see Table \ref{tab:results_overview}).
See Appendix Fig.~\ref{fig:calib_msp} for MSP reliability diagrams across all datasets. MSP is therefore well-suited for reliable ranking, but less effective when well-calibrated confidence scores are required.

\paragraph{Sample Consistency.} Sample Consistency, especially with separated decoding, offers low calibration error in several cases, such as TriviaQA ($ECE=0.134$), highlighting its advantage for confidence calibration. At the same time, its discrimination (AUROC) is generally lower across datasets compared to MSP, with the strongest values around $0.79$. Overall, Sample Consistency offers a favorable calibration–discrimination trade-off.

\begin{table}[htb]
\centering
\footnotesize
\setlength{\tabcolsep}{6pt}
\begin{adjustbox}{max width=\columnwidth}
\begin{tabular}{lcccc}
\toprule
\textbf{Dataset} & \textbf{Overall} & \textbf{Filtered} & \textbf{Improvement} & \textbf{Remaining} \\
& \textbf{ACC} & \textbf{ACC} & \textbf{(pp)} & \textbf{Samples} \\
\midrule
BoolQ     & 79.00\% & 90.85\% & +11.85 & 164 / 500 (32.8\%) \\
SQuAD     & 89.00\% & 95.98\% & +6.98  & 224 / 310 (72.3\%) \\
TriviaQA  & 60.80\% & 87.12\% & +26.32 & 264 / 500 (52.8\%) \\
GSM8K     & 68.00\% & 89.18\% & +21.18 & 194 / 500 (38.8\%) \\
\bottomrule
\end{tabular}
\end{adjustbox}
\caption{Selective prediction with CoCoA ($C_{CoCoA} > 0.8$). Accuracy improves substantially while keeping a large fraction of examples.}
\label{tab:cocoa_filtered}
\vspace{-0.5em}
\end{table}

\begin{figure}[h]
  \centering
  \includegraphics[width=\linewidth]{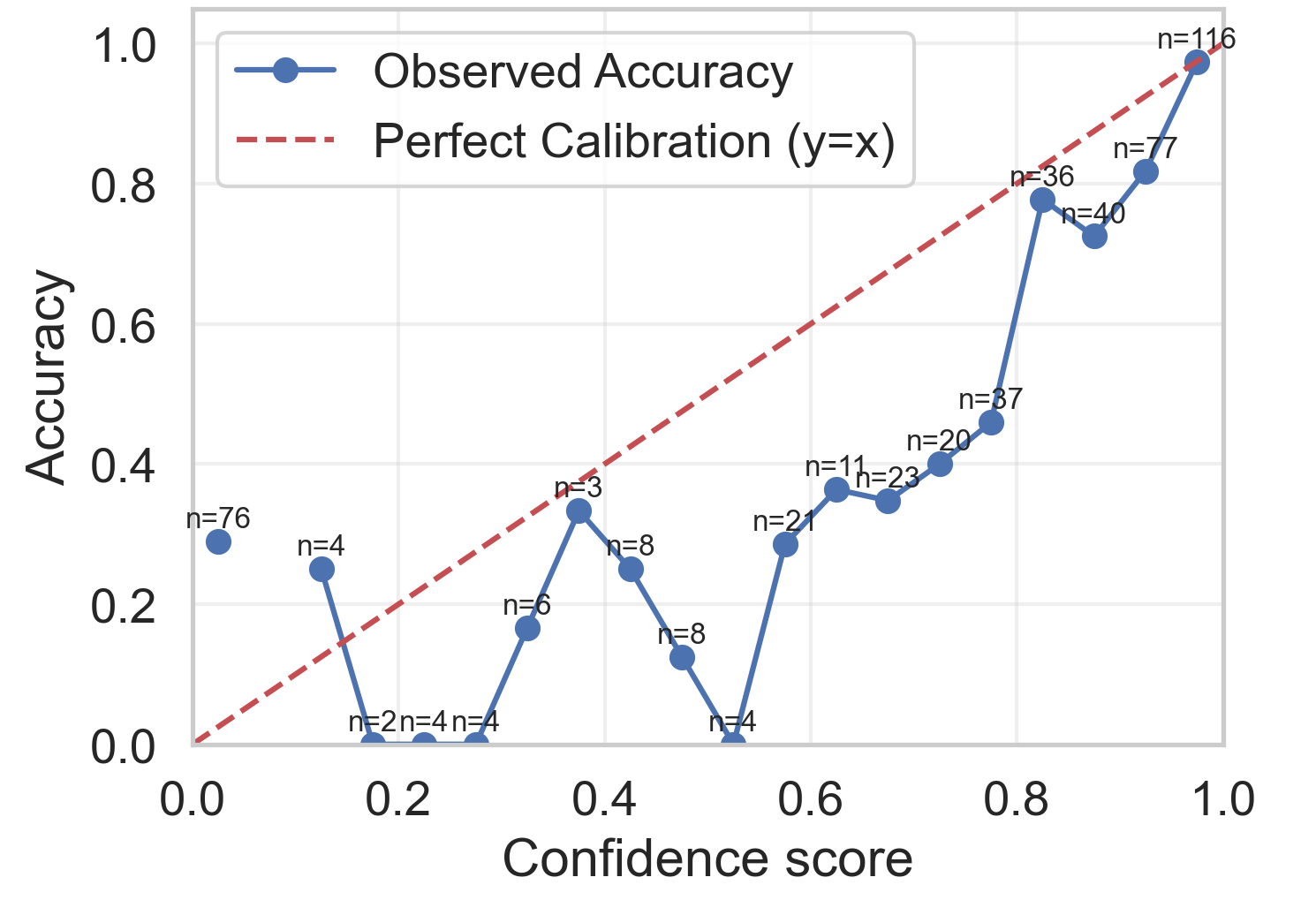}
  \caption{\textbf{Calibration of CoCoA on TriviaQA.} Bin-wise accuracy increases monotonically with confidence and aligns closely with the $y{=}x$ diagonal at high confidence, indicating well-calibrated scores in the decision-relevant range (bin sizes annotated).}
  \label{fig:cocoa_trivia_calibration}
\end{figure}

\paragraph{Selective prediction with \textbf{CoCoA}.}
A selective-prediction rule that accepts only high-confidence outputs can materially improve reliability.
Table~\ref{tab:cocoa_filtered} reports accuracy when \textbf{filtering by CoCoA confidence} $C_{CoCoA} > 0.8$ (absolute percentage-point gains and retained coverage shown). 
In short, \emph{precision rises sharply while retaining useful coverage}: gains are largest on harder tasks (TriviaQA, GSM8K), consistent with Figure~\ref{fig:cocoa_trivia_calibration} illustrates CoCoA on TriviaQA, where high-confidence bins closely match ground-truth accuracy.
See Appendix Fig.~\ref{fig:calib_all} for all CoCoA calibration plots on all datasets.

\section{Conclusion}
Task structure strongly shapes which uncertainty estimator is most useful. On the simplest setting-binary decisions in BoolQ-methods are closer together: MSP and CoCoA tie on accuracy while CoCoA provides the best ranking signal, and multi-sample VCE mostly reduces modest overconfidence. In contrast, on the most complex setting-multi, step reasoning in GSM8K, single-sample verbalized confidence is highly overconfident and weak at discrimination, while multi-sample aggregation markedly improves accuracy. The hybrid CoCoA yields the most calibrated scores (ECE $0.081$) even when MSP remains the strongest ranker. TriviaQA underscores the value of probability-based ranking under knowledge uncertainty. Importantly, for VCE we observe that TOP-K and SEP produce very similar reliability (average gaps $\sim0.03$ AUROC and $\sim0.015$ ECE), enabling practitioners to reuse alternatives from a single decoding to approximate separated decoding’s benefits and thereby save compute and latency. We reserve separated decoding for cases where small diversity gains matter. Temperature sweeps further show no universal best setting-optimal $T$ is task-specific-so moderate temperatures can help discrimination but do not by themselves fix miscalibration.\\[12pt]
Overall, our results suggest a practical recipe. MSP works best when ranking is most important, especially on knowledge-heavy tasks. Self-consistency helps to stabilize performance on difficult reasoning problems. CoCoA is preferable when downstream applications require calibrated and reliable confidence. When using verbalized confidence, TOP-K offers a good trade-off by reducing the cost of multi-sample decoding.

\section*{Limitations}
Our study focuses on question-answering tasks with relatively short outputs. The performance of uncertainty metrics may vary on other tasks such as long-form generation, dialogue, or code generation. 
Future work should evaluate whether the trends observed here hold for tasks with longer, more complex outputs. 
Additionally, our experiments used a single model (LLaMA~3.2 3B-Instruct). Larger models or different architectures might have different calibration characteristics. However, we expect the qualitative insights to generalize. 
Another limitation is the computational cost of multi-sample methods. Using TOP-K for VCE mitigates this, but Sample Consistency and CoCoA still require separated decodings. 
Finally, our evaluation assumes access to ground-truth correctness to measure calibration and AUROC.

\paragraph{Acknowledgment on AI Assistance.}

\href{https://www.deepl.com/en/home}{DeepL}
 and \href{https://chatgpt.com/}{ChatGPT}
 were used for translation, summarization, and grammatical or stylistic rephrasing of text. \href{https://github.com/features/copilot}{GitHub
 Copilot} was used to improve code efficiency and assist in generating auxiliary code components.

\paragraph{Acknowlegments}
This paper is supported by the DAAD programme Konrad Zuse Schools of Excellence in Artificial Intelligence, sponsored by the Federal Ministry of Research, Technology and Space.

\FloatBarrier

\bibliography{custom}

\FloatBarrier

\appendix
\renewcommand{\thesubsection}{A}
\setcounter{figure}{0}
\setcounter{table}{0}
\setcounter{footnote}{0}
\renewcommand\thefigure{A.\arabic{figure}}
\renewcommand\thetable{A.\arabic{table}}
\renewcommand\theHfigure{A.\arabic{figure}}
\renewcommand\theHtable{A.\arabic{table}}

\section{Additional Results and Figures}

\begin{figure*}
\centering
\begin{subfigure}[b]{0.49\textwidth}
\includegraphics[width=\linewidth]{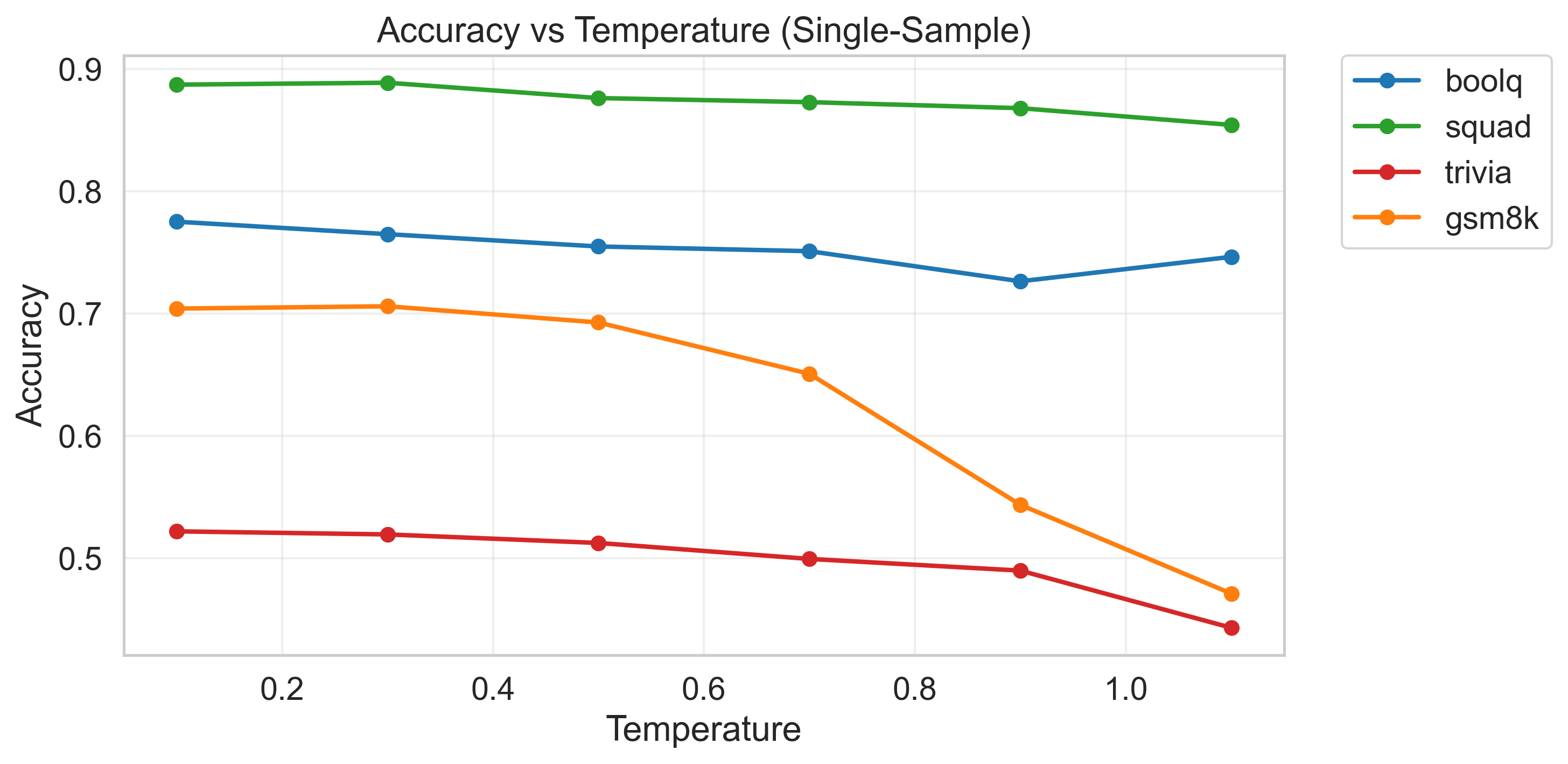}
\caption{Accuracy vs.\ temperature (single-sample VCE).}
\end{subfigure}\hfill
\begin{subfigure}[b]{0.49\textwidth}
\includegraphics[width=\linewidth]{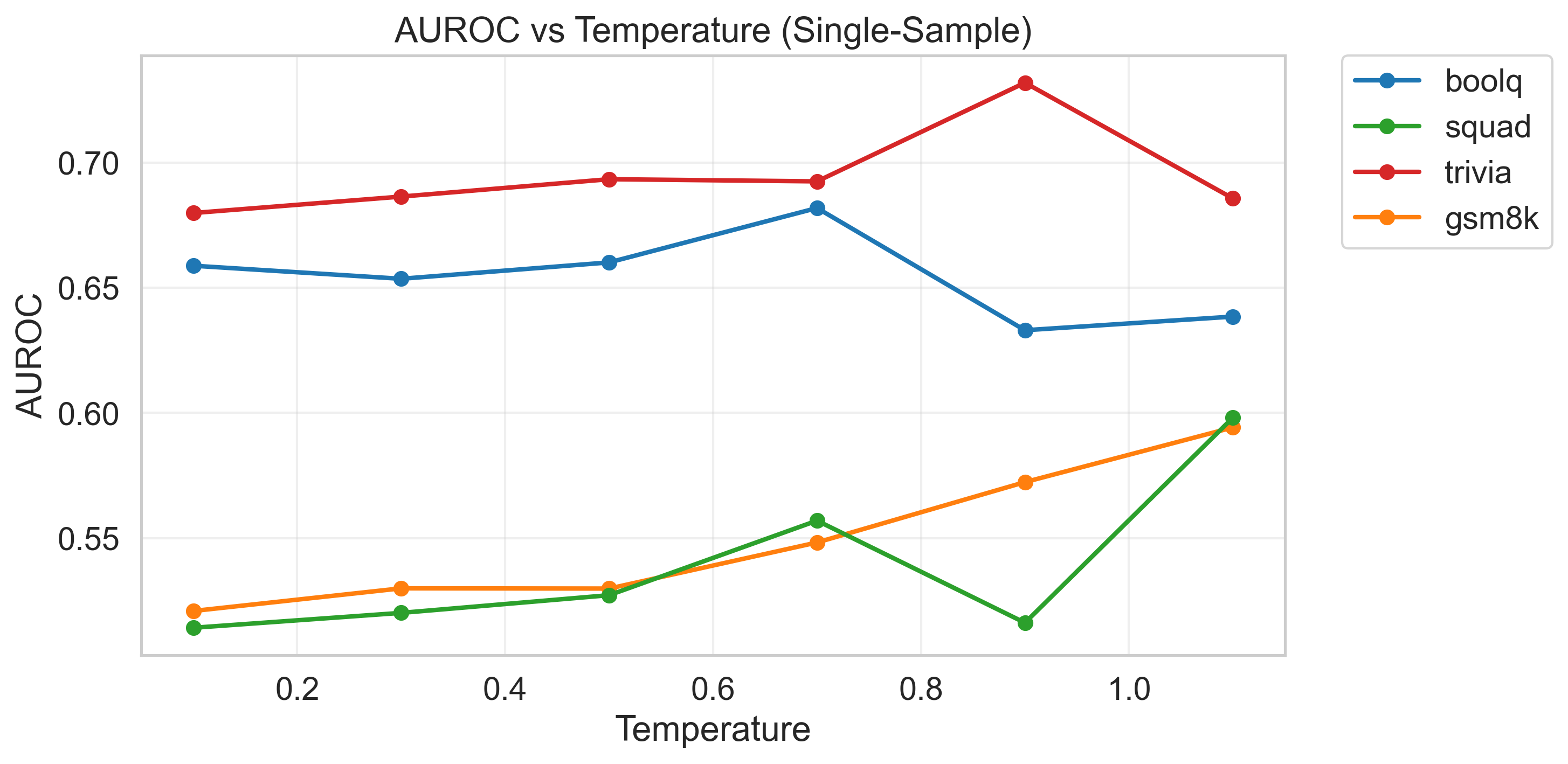}
\caption{AUROC vs.\ temperature (single-sample VCE).}
\end{subfigure}\\[0.5em]
\begin{subfigure}[b]{0.49\textwidth}
\includegraphics[width=\linewidth]{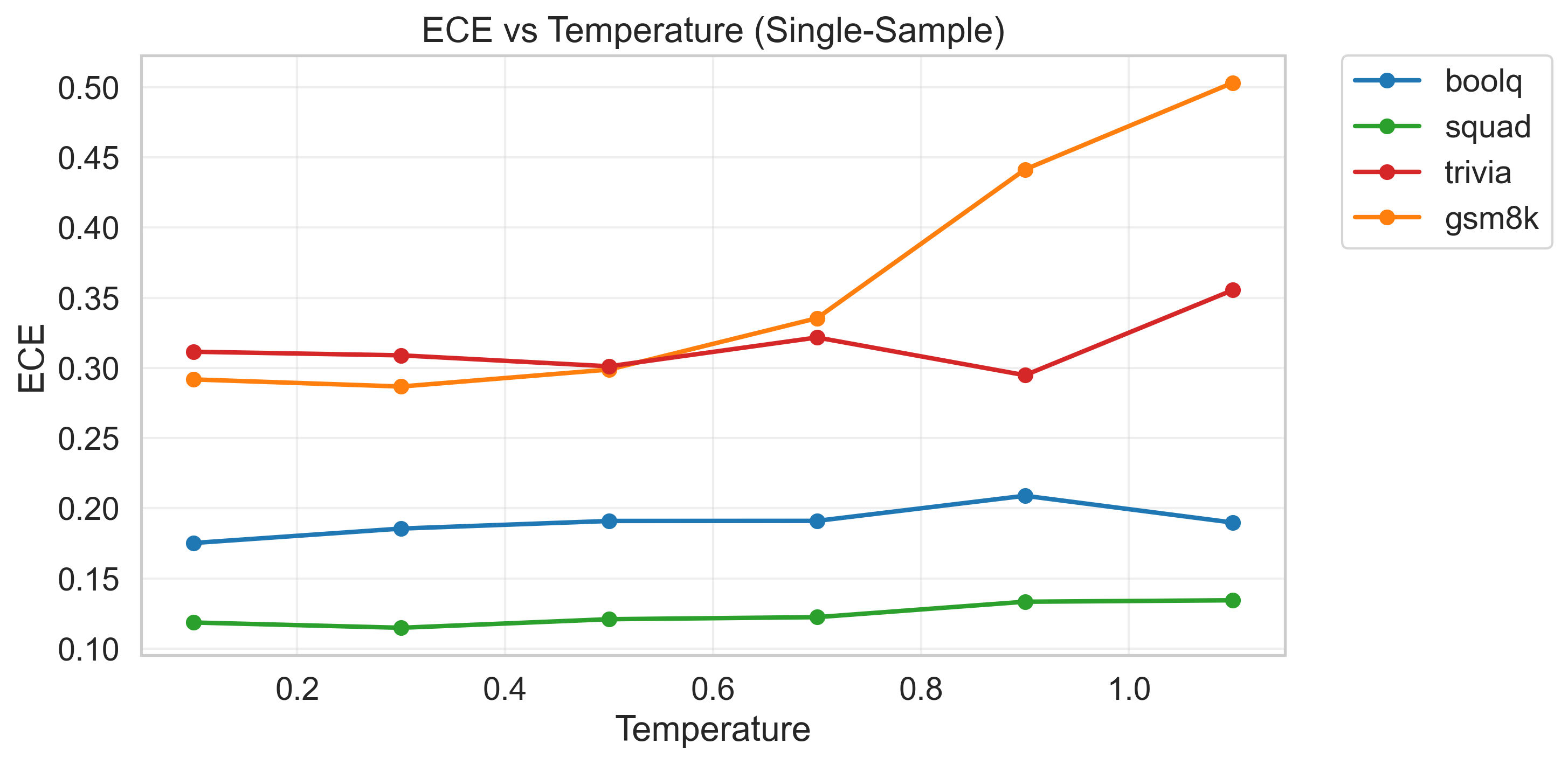}
\caption{ECE vs.\ temperature (single-sample VCE).}
\end{subfigure}\hfill
\begin{subfigure}[b]{0.49\textwidth}
\includegraphics[width=\linewidth]{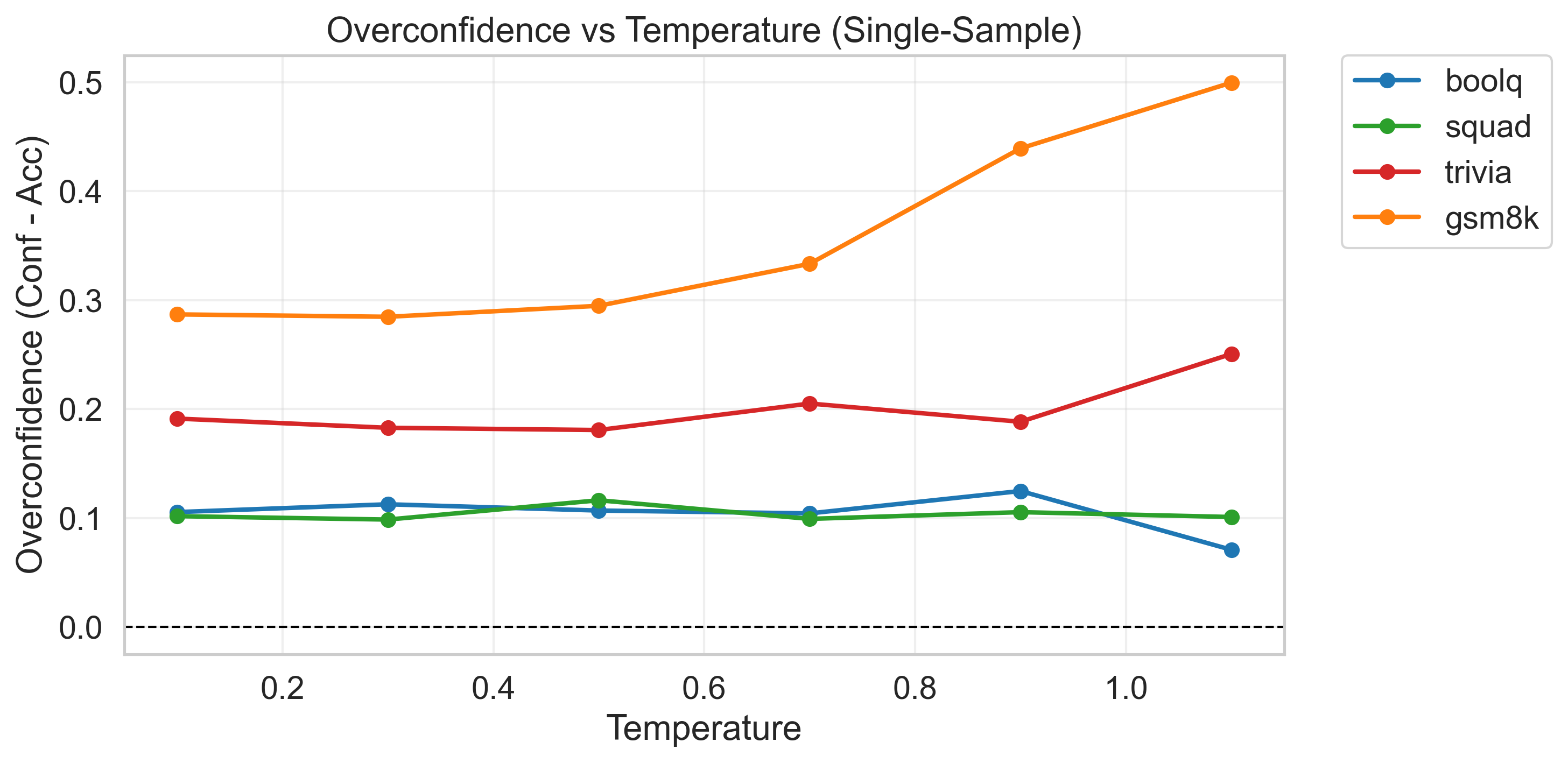}
\caption{Overconfidence vs.\ temperature (single-sample VCE).}
\end{subfigure}
\caption{Single-sample VCE temperature sweep. Optima are task-specific; no single $T$ dominates across datasets.}
\label{fig:temp_suite_single}
\end{figure*}

\begin{figure*}
\centering
\begin{subfigure}[b]{0.49\textwidth}
\includegraphics[width=\linewidth]{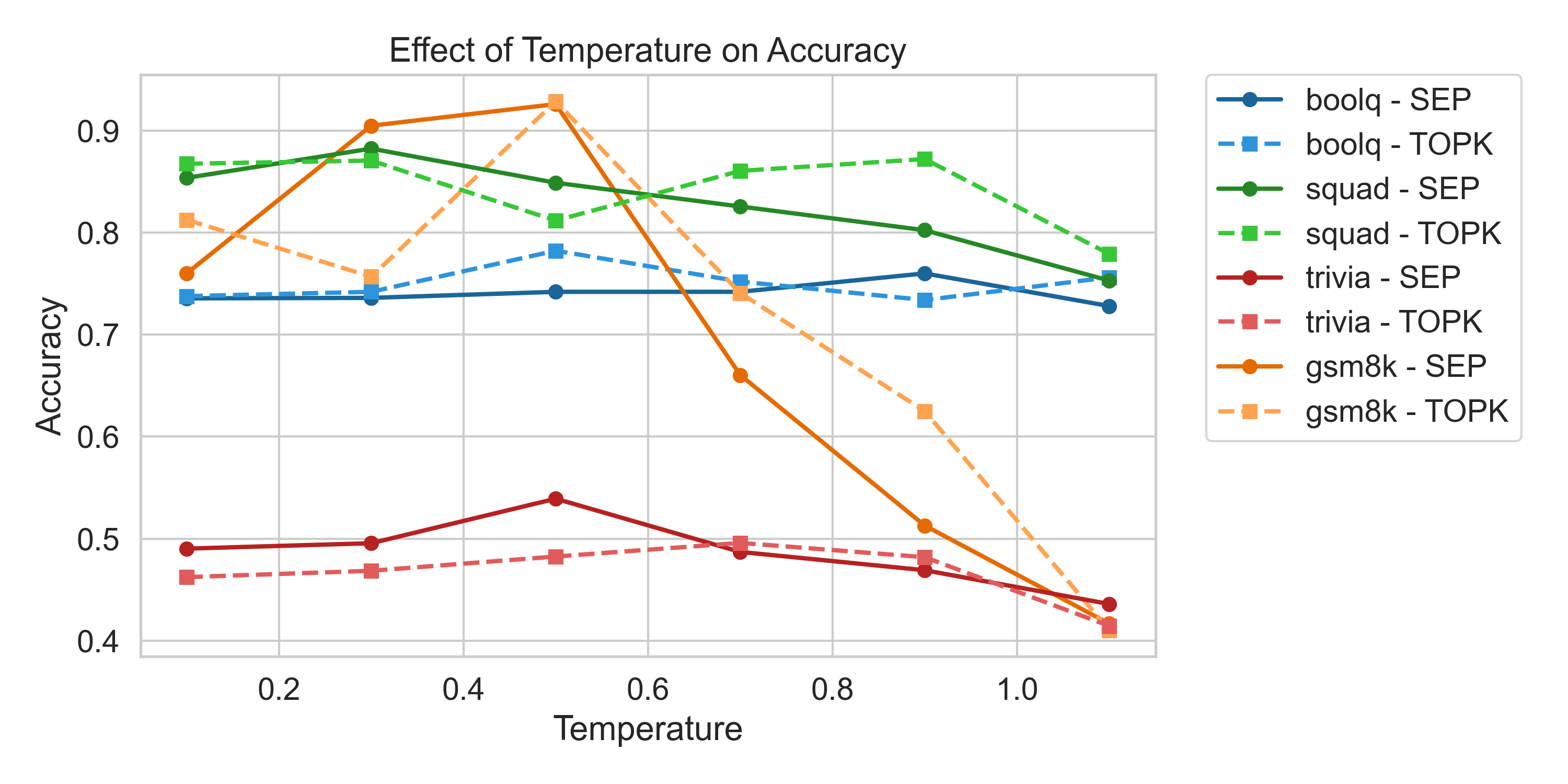}
\caption{Accuracy vs.\ temperature (multi-sample VCE, SEP vs.\ TOP-K).}
\end{subfigure}\hfill
\begin{subfigure}[b]{0.49\textwidth}
\includegraphics[width=\linewidth]{multisample_AUROC.png}
\caption{AUROC vs.\ temperature (multi-sample VCE, SEP vs.\ TOP-K).}
\end{subfigure}\\[0.5em]
\begin{subfigure}[b]{0.49\textwidth}
\includegraphics[width=\linewidth]{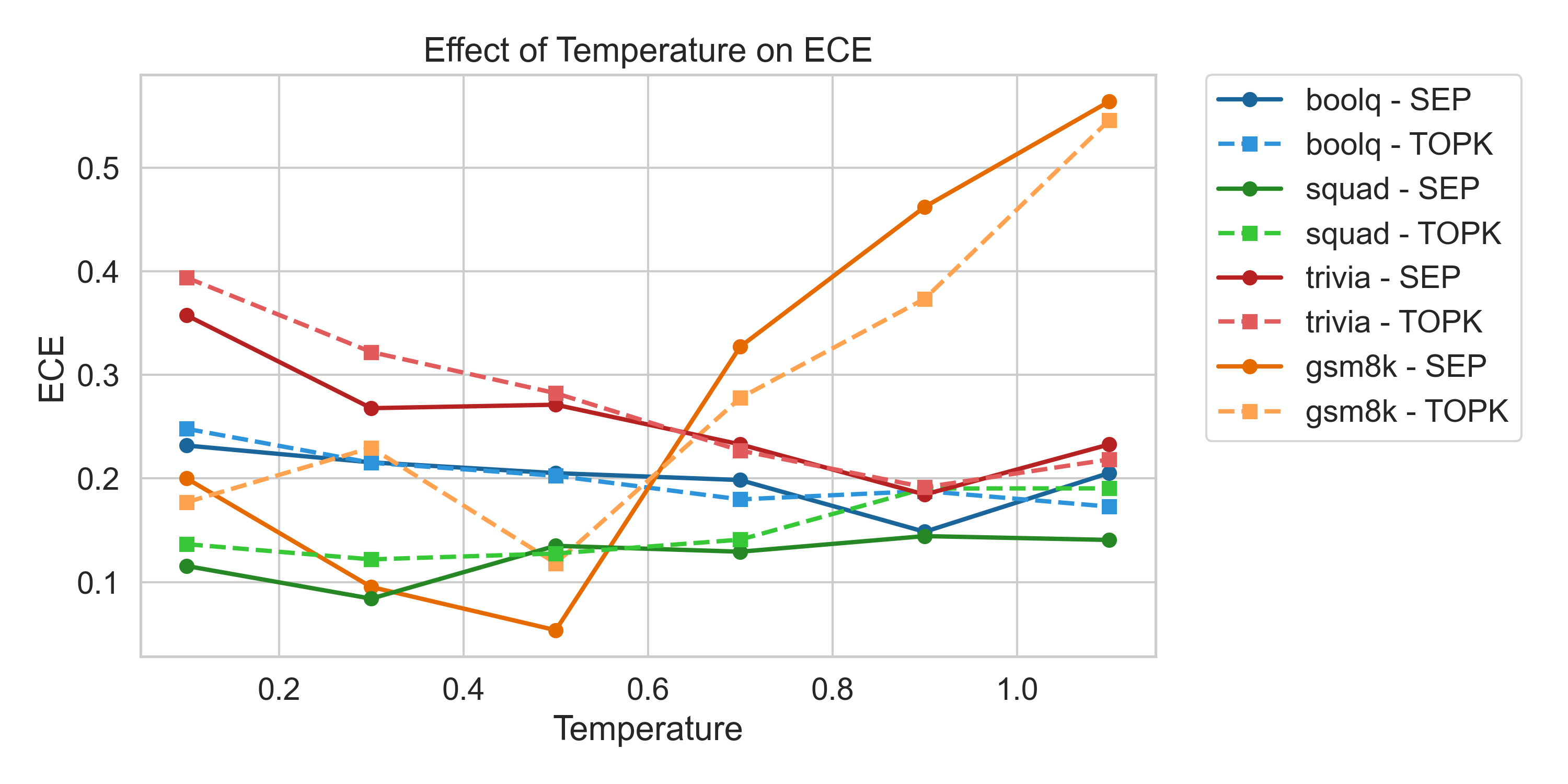}
\caption{ECE vs.\ temperature (multi-sample VCE, SEP vs.\ TOP-K).}
\end{subfigure}\hfill
\begin{subfigure}[b]{0.49\textwidth}
\includegraphics[width=\linewidth]{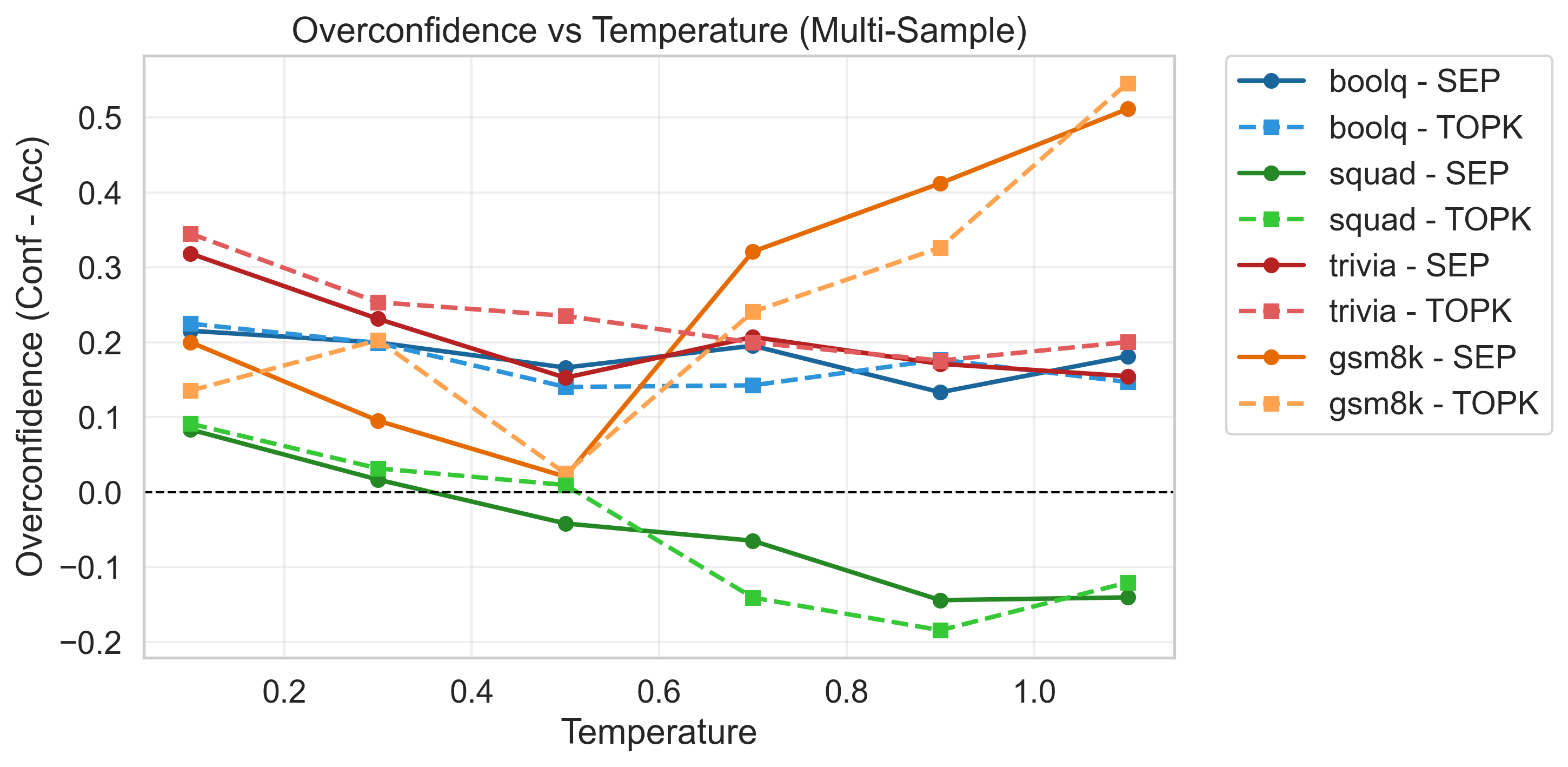}
\caption{Overconfidence vs.\ temperature (multi-sample VCE, SEP vs.\ TOP-K).}
\end{subfigure}
\caption{Multi-sample VCE temperature sweep. Optima are task-specific; no single $T$ dominates across datasets; SEP and TOP-K are broadly comparable.}
\label{fig:temp_suite_multi}
\end{figure*}

\FloatBarrier

\captionsetup[subfigure]{justification=raggedright,singlelinecheck=false,font=small}

\begin{figure*}[p]
  \centering
  \begin{subfigure}{.9\linewidth}
    \includegraphics[width=\linewidth,height=.2\textheight,keepaspectratio]{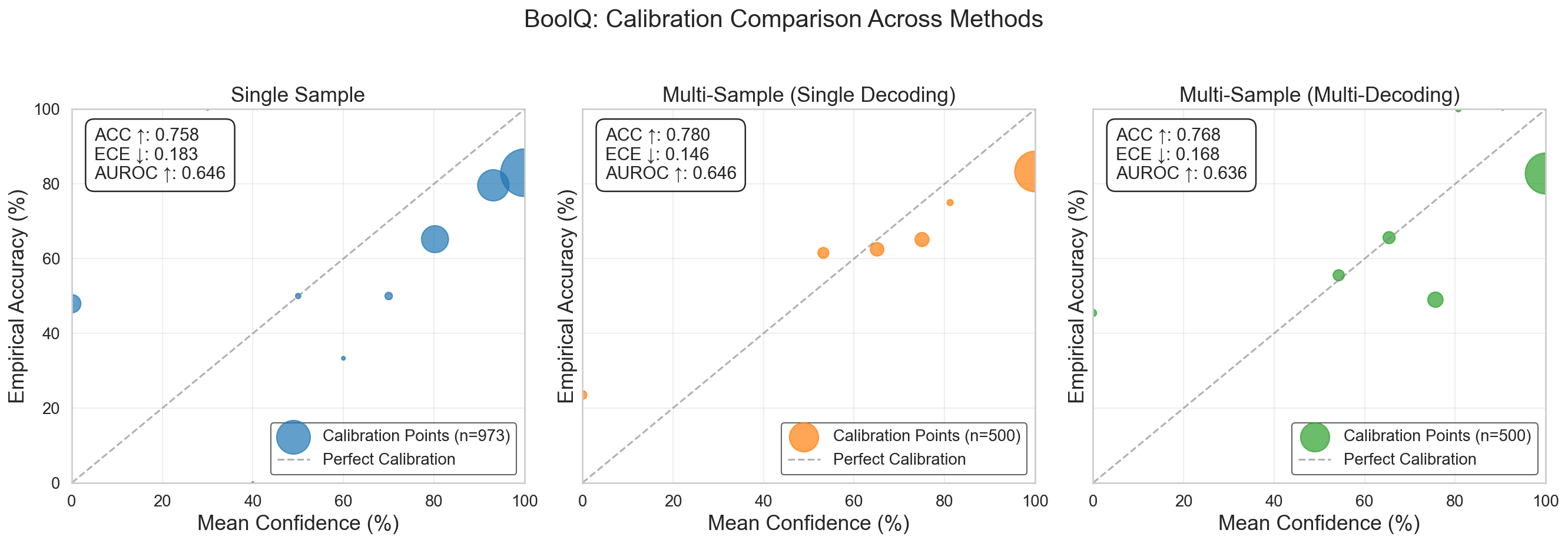}
    \caption{\textbf{BoolQ.} Multi-sample corrects overconfident high bins; AUROC is stable.}
  \end{subfigure}\vspace{0.6ex}

  \begin{subfigure}{.9\linewidth}
    \includegraphics[width=\linewidth,height=.2\textheight,keepaspectratio]{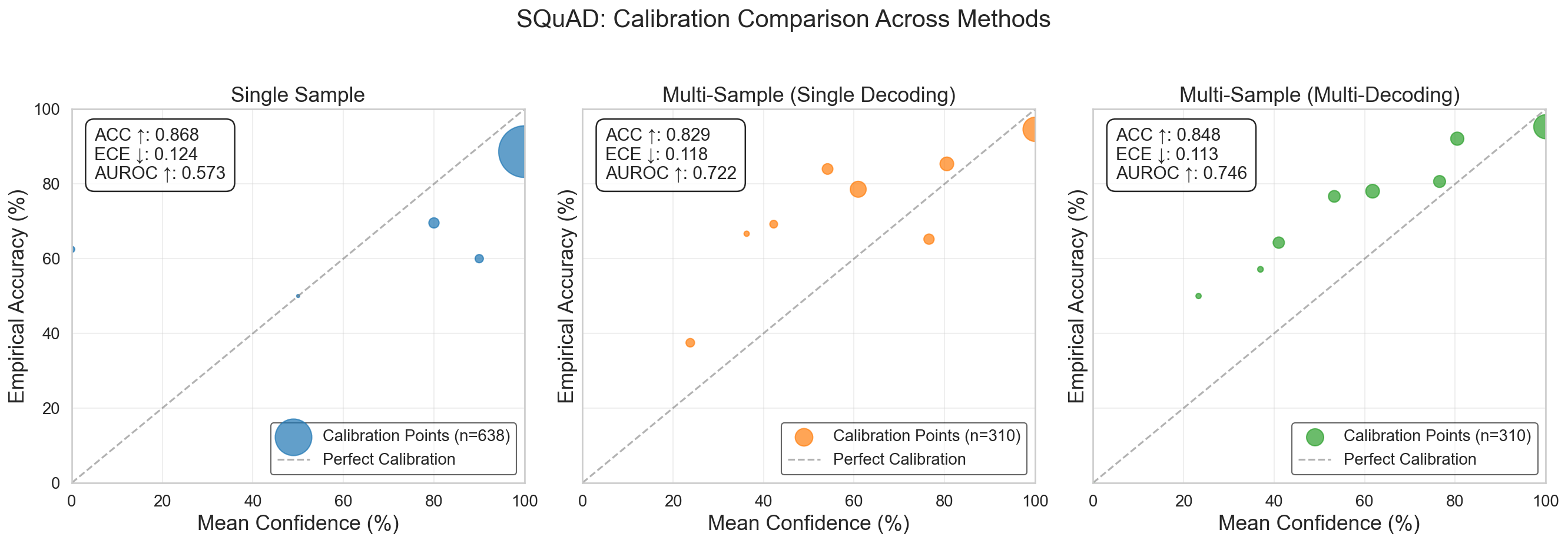}
    \caption{\textbf{SQuAD\,2.0.} AUROC improves strongly; ECE reduces to \(\mathbf{0.113}\) (SEP).}
  \end{subfigure}\vspace{0.6ex}

  \begin{subfigure}{.9\linewidth}
    \includegraphics[width=\linewidth,height=.2\textheight,keepaspectratio]{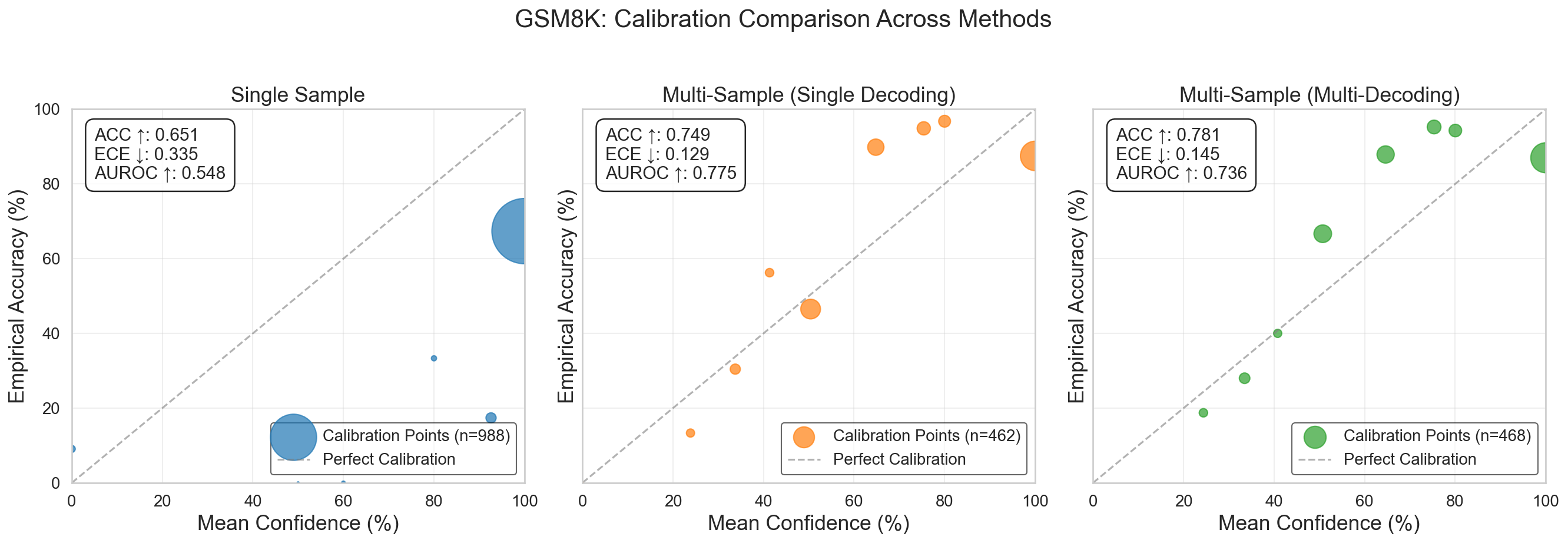}
    \caption{\textbf{GSM8K.} Largest relative ECE drop; ACC and AUROC improve substantially.}
  \end{subfigure}\vspace{0.6ex}

  \begin{subfigure}{.9\linewidth}
    \includegraphics[width=\linewidth,height=.2\textheight,keepaspectratio]{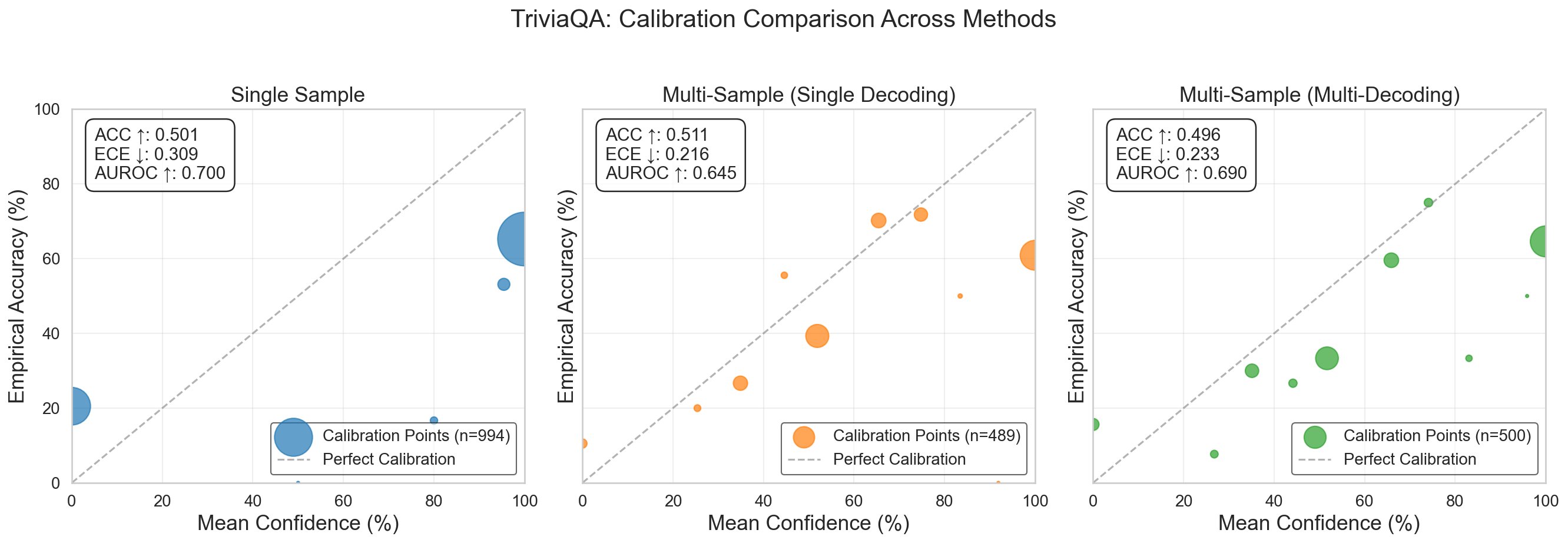}
    \caption{\textbf{TriviaQA (open-domain).} Large ECE drop; ACC comparable and AUROC slightly lower than single-sample.}
  \end{subfigure}

  \caption{\textbf{VCE: calibration by decoding method.} Points are confidence bins (bubble size = bin count); dashed line is perfect calibration \(y{=}x\).}
  \label{fig:vce_methods}
\end{figure*}

\begin{figure*}[ht!]
    \centering
    \begin{minipage}{0.45\textwidth}
        \centering
        \includegraphics[width=\linewidth]{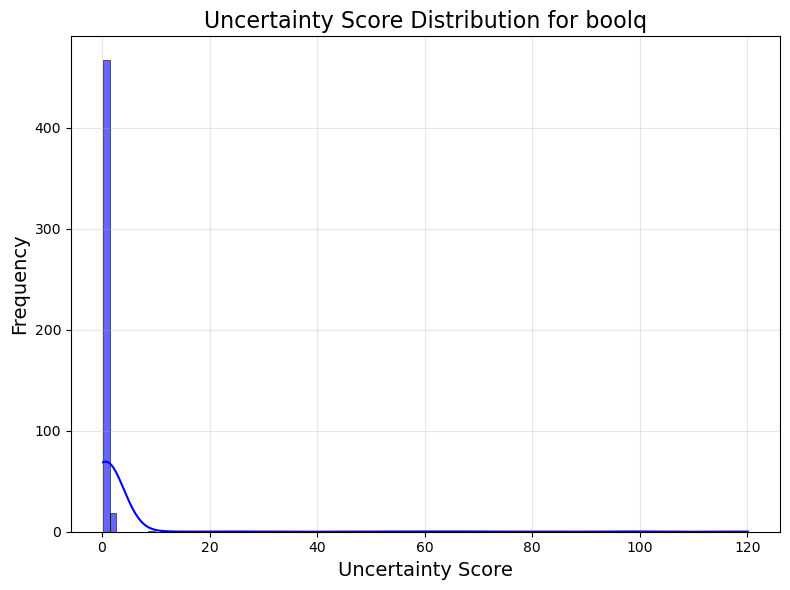}
    \end{minipage}
    \hfill
    \begin{minipage}{0.45\textwidth}
        \centering
        \includegraphics[width=\linewidth]{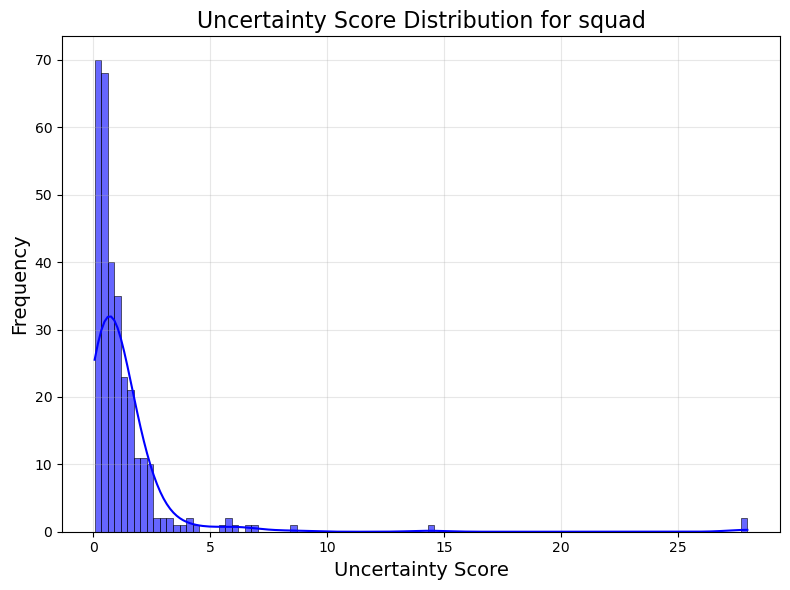}
    \end{minipage}
    \vspace{0.5cm}
    \begin{minipage}{0.45\textwidth}
        \centering
        \includegraphics[width=\linewidth]{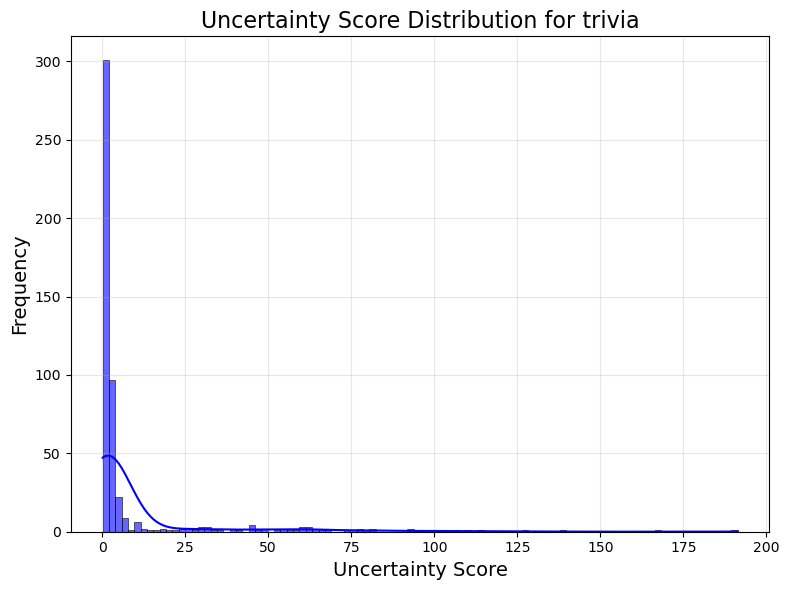}
    \end{minipage}
    \hfill
    \begin{minipage}{0.45\textwidth}
        \centering
        \includegraphics[width=\linewidth]{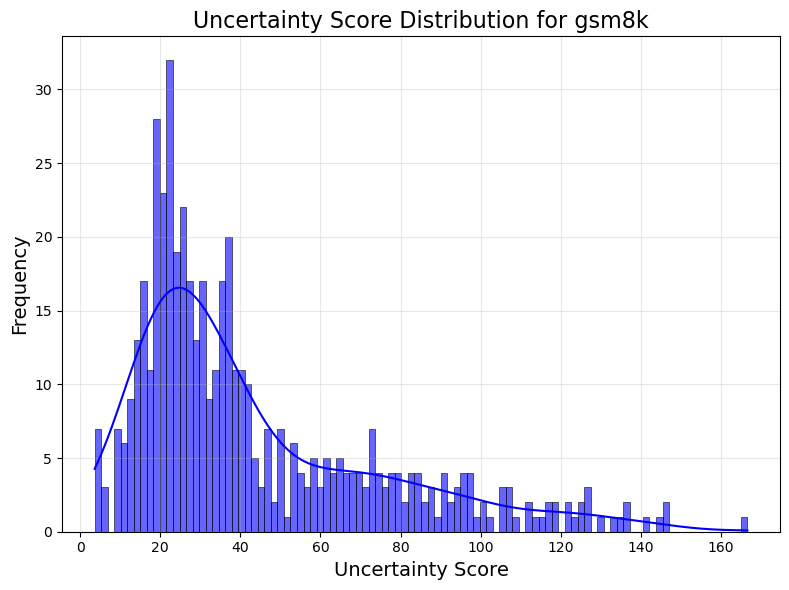}
    \end{minipage}
    \caption{Distribution of raw MSP uncertainty scores}
    \label{fig:msp_uncertainty}
\end{figure*}

\begin{figure*}[t]
\centering
\begin{subfigure}[b]{.8\linewidth}
\includegraphics[width=\linewidth]{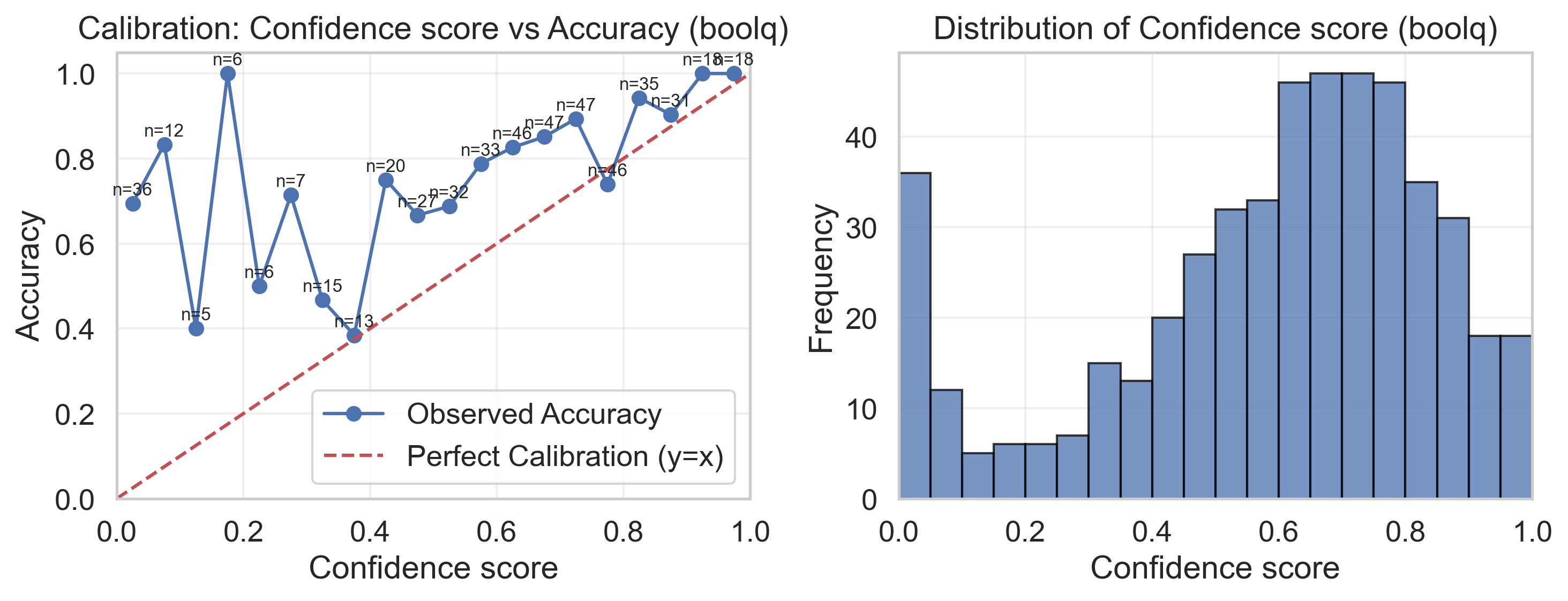}
\caption{BoolQ}
\end{subfigure}\hfill
\begin{subfigure}[b]{.8\linewidth}
\includegraphics[width=\linewidth]{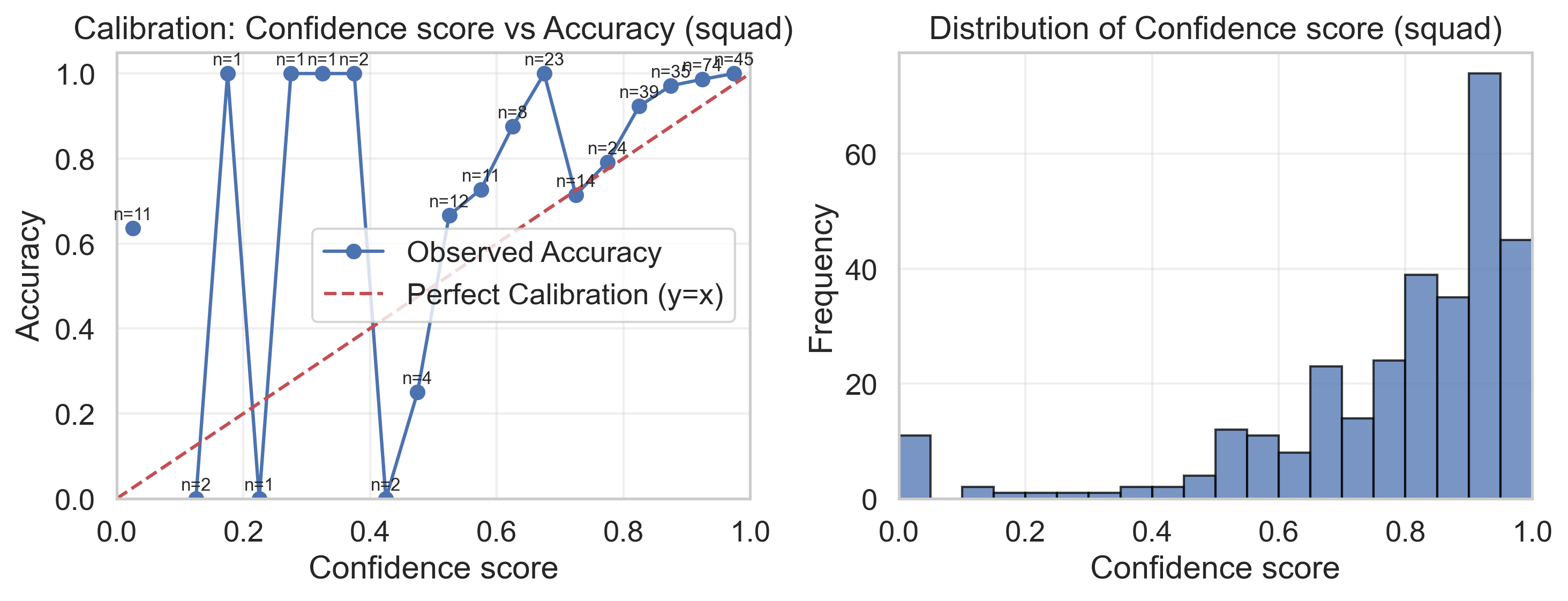}
\caption{SQuAD\,2.0}
\end{subfigure}\\
\begin{subfigure}[b]{.8\linewidth}
\includegraphics[width=\linewidth]{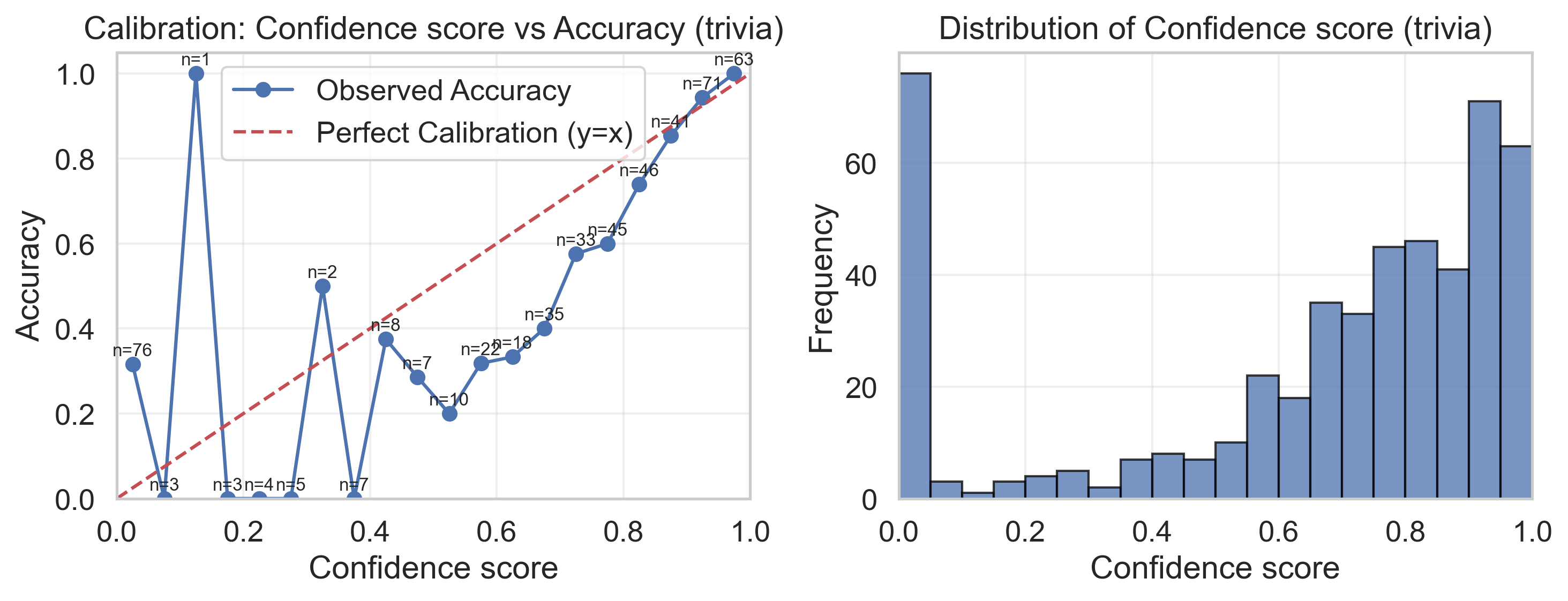}
\caption{TriviaQA}
\end{subfigure}\hfill
\begin{subfigure}[b]{0.8\textwidth}
\includegraphics[width=\linewidth]{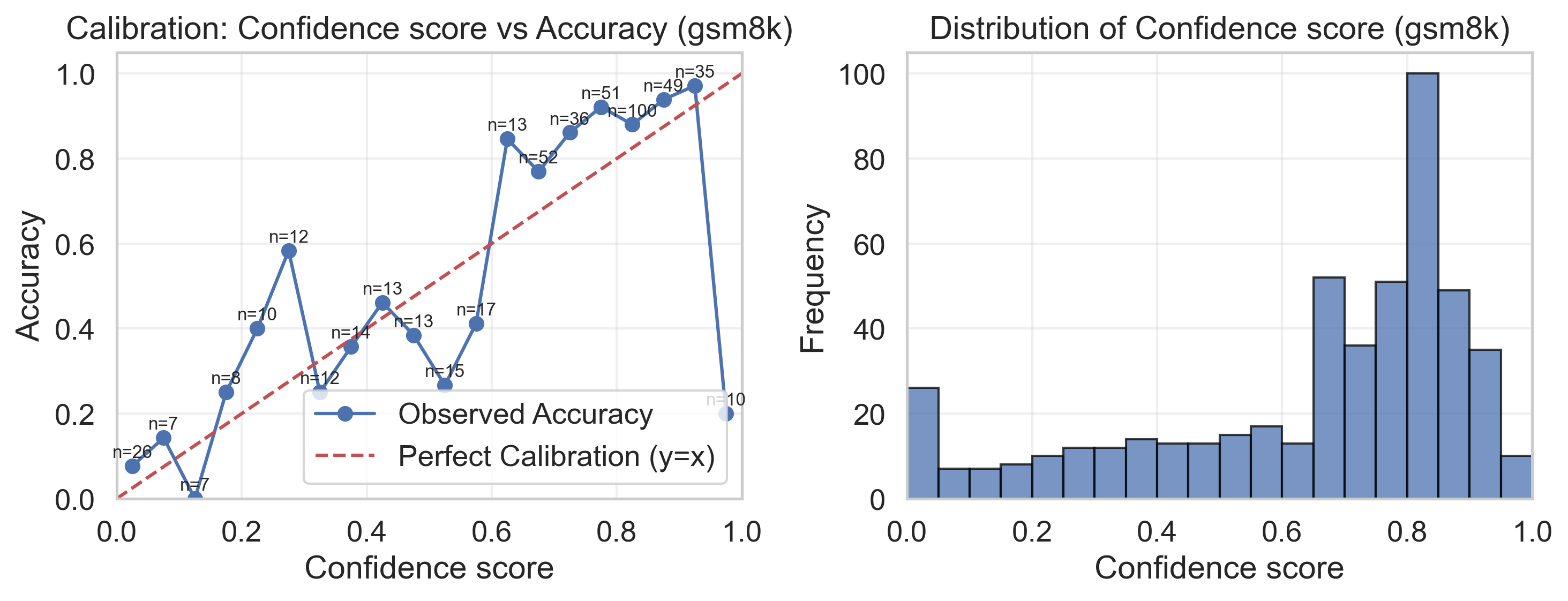}
\caption{GSM8K}
\end{subfigure}
\caption{\textbf{MSP calibration with $C_{MSP}$.} Reliability (left) and confidence distribution (right). Labels above points denote bin counts. MSP aligns high-confidence bins with $y{=}x$ on SQuAD, TriviaQA and BoolQ.}
\label{fig:calib_msp}
\end{figure*}

\begin{figure*}[t]
\centering
\begin{subfigure}[b]{.8\linewidth}
\includegraphics[width=\linewidth]{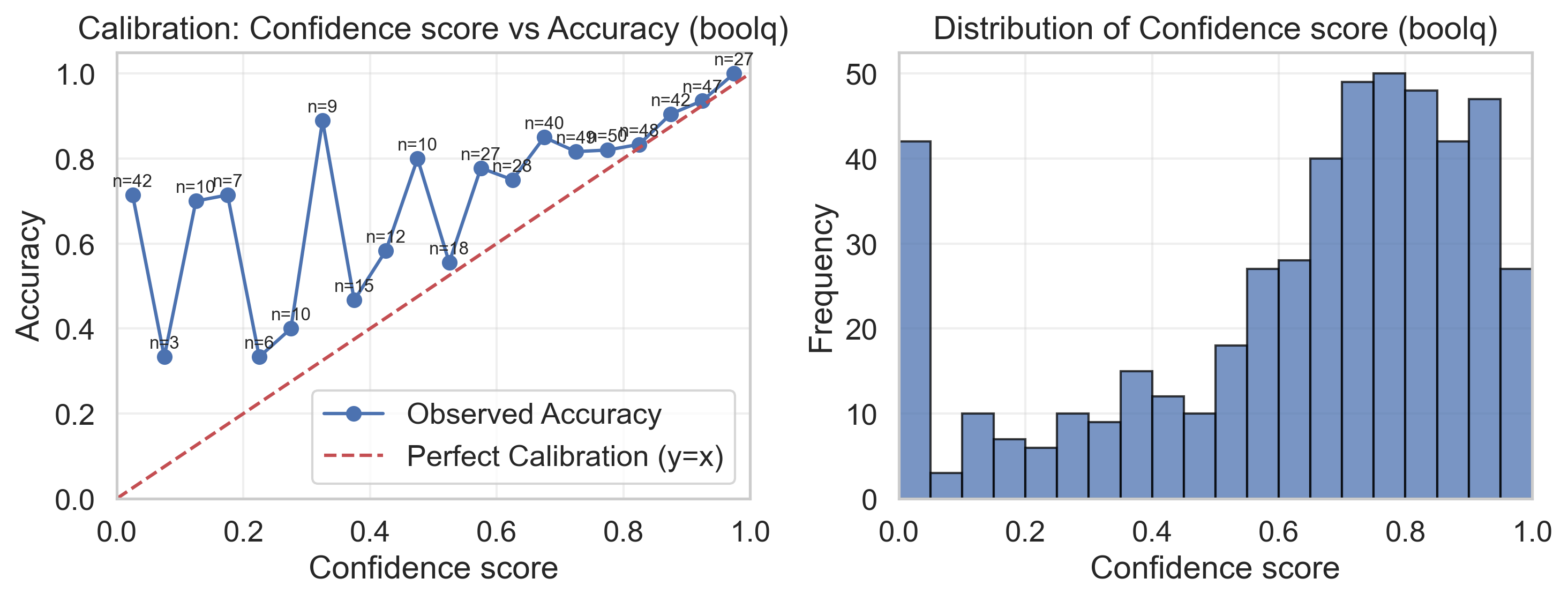}
\caption{BoolQ}
\end{subfigure}\hfill
\begin{subfigure}[b]{.8\linewidth}
\includegraphics[width=\linewidth]{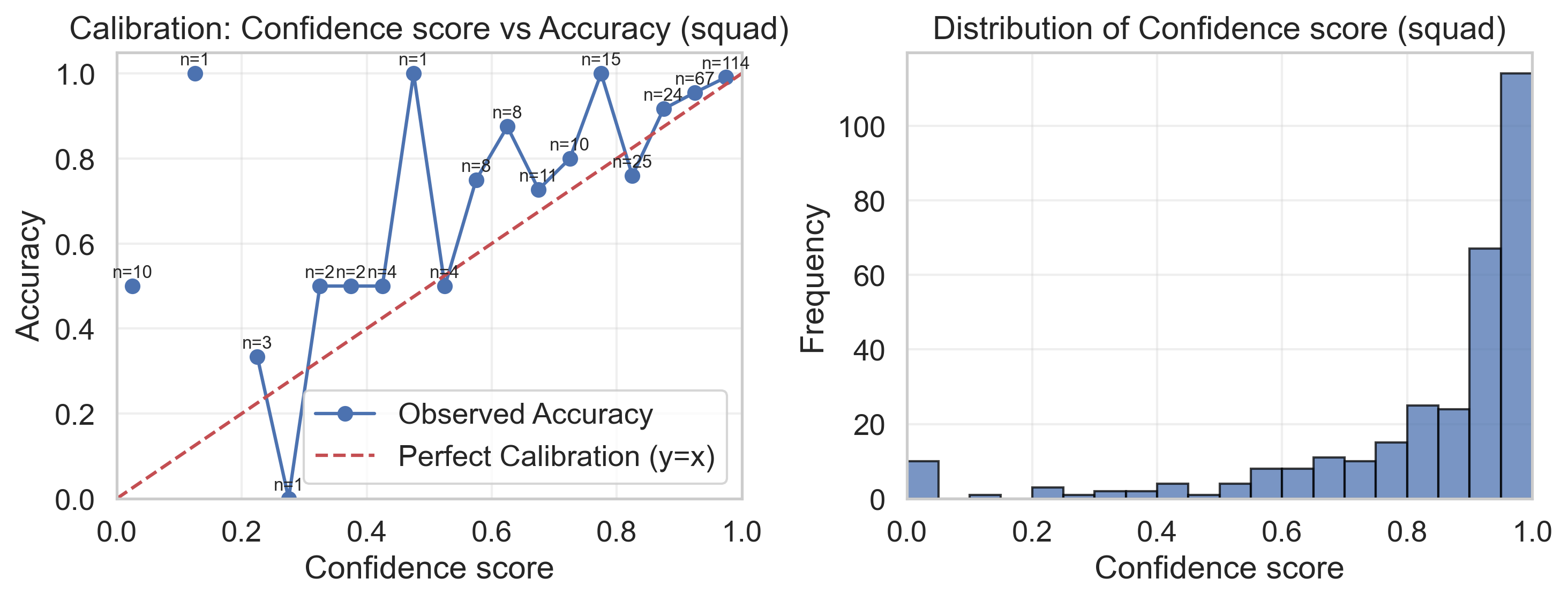}
\caption{SQuAD\,2.0}
\end{subfigure}\\
\begin{subfigure}[b]{.8\linewidth}
\includegraphics[width=\linewidth]{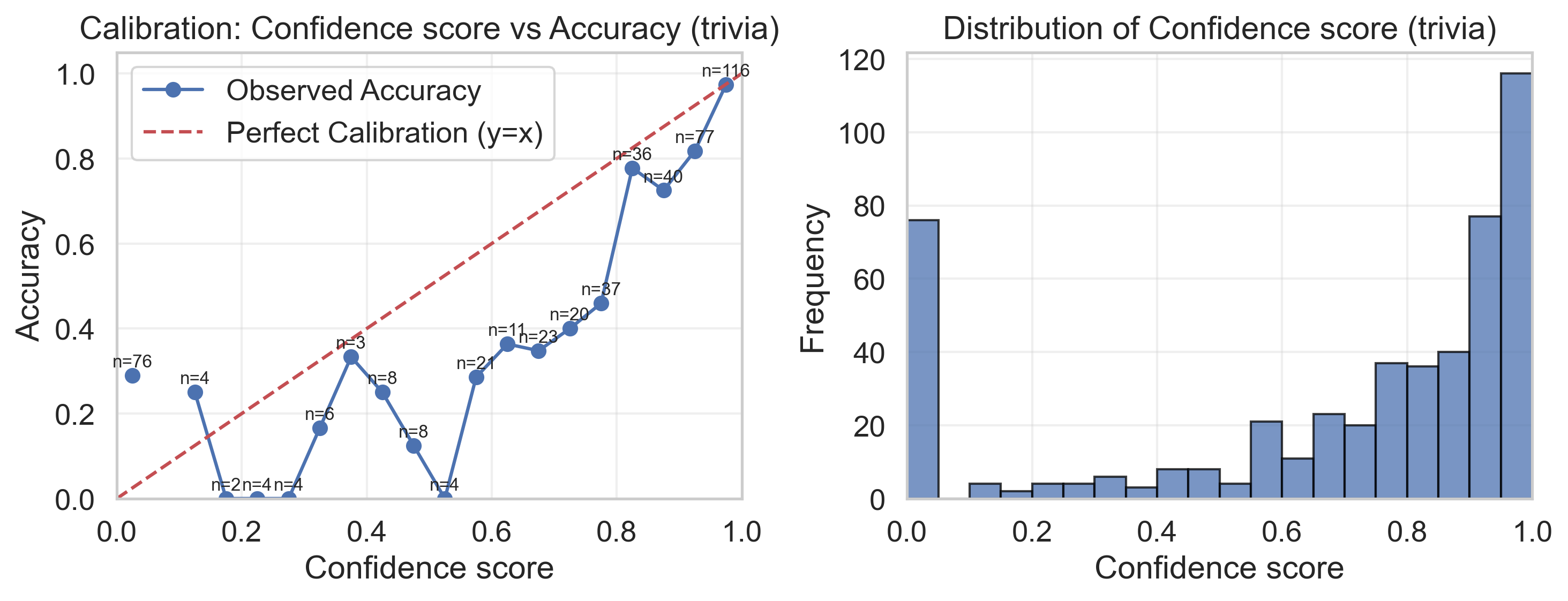}
\caption{TriviaQA}
\end{subfigure}\hfill
\begin{subfigure}[b]{0.8\textwidth}
\includegraphics[width=\linewidth]{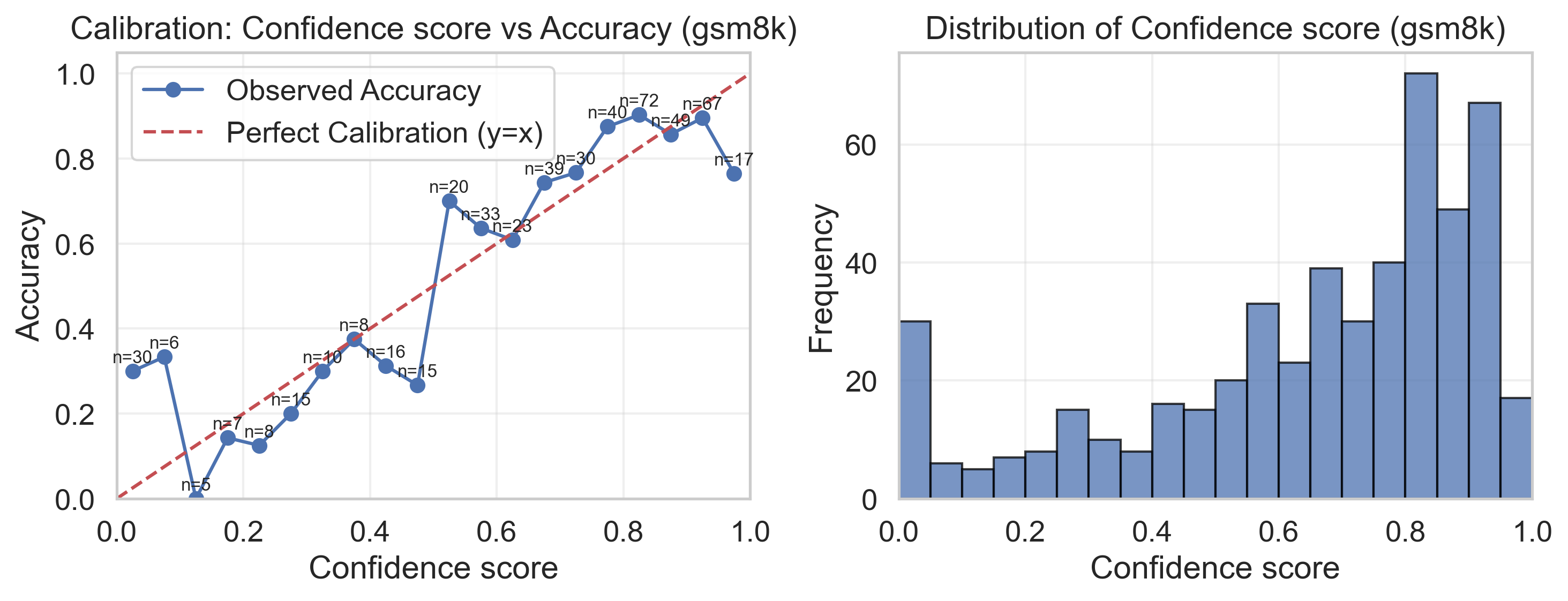}
\caption{GSM8K}
\end{subfigure}
\caption{\textbf{CoCoA calibration with $C_{CoCoA}$.} Reliability (left) and confidence distribution (right). Labels above points denote bin counts. CoCoA aligns high-confidence bins with $y{=}x$ on SQuAD/TriviaQA and matches BoolQ’s best VCE calibration, and tracks the diagonal well on GSM8K.}
\label{fig:calib_all}
\end{figure*}

\end{document}